%% file: main.tex
\documentclass[journal]{IEEEtran}

\input{sec/0_preamble.tex}

\PassOptionsToPackage{natbib=true}{biblatex}
\usepackage[style=numeric,backend=biber,language=english,sorting=none,maxbibnames=99,doi=false]{biblatex}
\usepackage{cleveref}
\begin{document}

\title{Event Detection in Videos: A Framework for the Development of New Methods}

\author{Anastasia Zakharova, Thierry Bouwmans, Anthony Cioppa, Adrien Deli{\`e}ge, Antonio Greco, Ana{\"i}s Halin, Kamil Jeziorek, Meghna Kapoor, Tomasz Kryjak, Islam Osman, S{\'e}bastien Pi{\'e}rard, Carlo Sansone, Mohamed S. Shehata, Renaud Vandeghen, Marc Van Droogenbroeck, Bruno Vento,~\IEEEmembership{Staff,~IEEE,}
\thanks{Anastasia Zakharova and Thierry Bouwmans  are with the MIA lab of the La Rochelle University, France.}%
\thanks{Antonio Greco is with DIEM Department  of the University of Salerno, Italy.}%
\thanks{Kamil Jeziorek and Tomasz Kryjak are  with the AGH University of Krakow, Poland}%
\thanks{Meghna Kapoor is  with the L3i lab of the La Rochelle University, France.}%
\thanks{Islam Osman  and  Mohamed S Shehata are  with the Department of Computer Science, The University of British Columbia, Kelowna, Canada .}%
\thanks{Anthony Cioppa, Adrien Deli{\`e}ge, Ana{\"i}s Halin, S{\'e}bastien Pi{\'e}rard, Renaud Vandeghen, and  Marc Van Droogenbroeck are with the Montefiore Institute of the University of Li{\`e}ge, Belgium.}%
\thanks{Carlo Sansone is with DIETI Department University of Napoli Federico II, Italy.}%
\thanks{Bruno Vento is with Consorzio Interuniversitario Nazionale per l'Informatica (CINI), Italy.}
}

\maketitle

\begin{abstract}
Event detection tasks in videos, the most important aspect of video surveillance, aim to detect events either at the pixel-level, frame-level, or clip-level. Plenty of methods intended for event detection in different environments, for various applications, and within different acquisition techniques were introduced. Naturally, the attempts were made as well to classify these algorithms in terms of detection of performance or in terms of real-time abilities. Nevertheless, the lack of a large-scale dataset as well as rigorous performance evaluation methods have biased such comparisons as well as the development of the methods. 

Given the diversity of existing approaches, we believe it is essential for researchers to position their work within such a rich landscape. Thus, we propose a rigorous framework for developing new methods in event detection for videos.
Specifically, this framework is based on three main pillars: datasets, performance evaluation, and scenarios for deploying methods.
\end{abstract}


\input{sec/1_introduction}

\input{sec/2_related_work_new}

\input{sec/3_dataset}

\input{sec/4_performance}

\input{sec/5_scenarios}

\input{sec/7_conclusion}

\input{sec/9_acknowledgment}

\printbibliography

\input{sec/10_appendix_data}

\end{document}

%% file: sec/0_preamble.tex
\usepackage{amsmath,amsfonts}
\usepackage{algorithmic}
\usepackage{algorithm}
\usepackage{array}
\usepackage[caption=false,font=normalsize,labelfont=sf,textfont=sf]{subfig}
\usepackage{textcomp}
\usepackage{stfloats}
\usepackage{xurl}
\usepackage{verbatim}
\usepackage{graphicx}
\usepackage{xcolor}
\definecolor{cvprblue}{rgb}{0.21,0.49,0.74}
\usepackage[breaklinks,colorlinks,allcolors=cvprblue]{hyperref}

\usepackage[capitalize]{cleveref}

\usepackage{tikz}
\usepackage{xspace}
\usepackage{booktabs}
\usepackage{multirow}
\usepackage{tabularx}
\usepackage{array}
\def\CDNET{CDnet\xspace}
\def\fsd{Foreground Segmentation Dataset\xspace}
\def\fsdAcronym{FSD\xspace}
\def\datasetAnaisName{Synthetic Urban Crossroad\xspace}
\def\datasetAnaisAcronym{SUC\xspace}

\makeatletter
\DeclareRobustCommand\onedot{\futurelet\@let@token\@onedot}
\def\@onedot{\ifx\@let@token.\else.\null\fi\xspace}

\def\aka{\emph{a.k.a}\onedot} 
\def\eg{\emph{e.g}\onedot} 
\def\ie{\emph{i.e}\onedot} 
 
\def\etc{\emph{etc}\onedot} 
\def\wrt{w.r.t\onedot} 

\makeatother

\usepackage{amsthm}
\makeatletter
\newtheoremstyle{experimentstyle}%
  {0pt}{0pt}%
  {\normalfont}%
  {}%
  {\bfseries}%
  {}%
  {.5em}%
  {\thmname{#1}\thmnumber{ #2}{\normalfont:}\thmnote{ {\normalfont\itshape#3.}}}%
\makeatother
\theoremstyle{experimentstyle}
\newtheorem{experiment}{Random Experiment}

\newcommand{\outTN}{\ensuremath{tn}\xspace}%
\newcommand{\outFP}{\ensuremath{fp}\xspace}%
\newcommand{\outFN}{\ensuremath{fn}\xspace}%
\newcommand{\outTP}{\ensuremath{tp}\xspace}%
\newcommand{\TN}{\ensuremath{\text{TN}}\xspace}%
\newcommand{\FP}{\ensuremath{\text{FP}}\xspace}%
\newcommand{\FN}{\ensuremath{\text{FN}}\xspace}%
\newcommand{\TP}{\ensuremath{\text{TP}}\xspace}%

\newcommand{\pprior}{\ensuremath{\pi^{+}}\xspace}%
\newcommand{\nprior}{\ensuremath{\pi^{-}}\xspace}%
\newcommand{\prate}{\ensuremath{\tau^{+}}\xspace}%
\newcommand{\nrate}{\ensuremath{\tau^{-}}\xspace}%
\newcommand{\precision}{\ensuremath{\mathit{Pr}}\xspace}%
\newcommand{\recall}{\ensuremath{\mathit{Re}}\xspace}%
\newcommand{\tpr}{\ensuremath{\mathit{TPR}}\xspace}%
\newcommand{\fpr}{\ensuremath{\mathit{FPR}}\xspace}%
\newcommand{\tnr}{\ensuremath{\mathit{TNR}}\xspace}%
\newcommand{\ppv}{\ensuremath{\mathit{PPV}}\xspace}%
\newcommand{\npv}{\ensuremath{\mathit{NPV}}\xspace}%

\newcommand{\accuracy}{\ensuremath{A}\xspace}%
\newcommand{\iou}{\ensuremath{\mathit{IoU}}\xspace}%
\newcommand{\ba}{\ensuremath{\mathit{BA}}\xspace}%
\newcommand{\fone}{\ensuremath{F_1}\xspace}%
\newcommand{\fbeta}{\ensuremath{F_\beta}\xspace}%
\newcommand{\TileA}{\ensuremath{a}\xspace}%
\newcommand{\TileB}{\ensuremath{b}\xspace}%

\newcommand{\MEM}{\ensuremath{\mbox{MEM}}\xspace}%
\newcommand{\delay}{\ensuremath{\mbox{D}}\xspace}%
\newcommand{\PFR}{\ensuremath{\mbox{PFR}}\xspace}%

\global\long\def\comma{\,,}%
\global\long\def\point{\,.}%

\newcommand{\flops}{FLOPs\xspace}

%% file: sec/1_introduction.tex
\section{Introduction}
\label{sec:introduction}

\IEEEPARstart{T}{he} event detection task is a crucial task in computer vision since its origin, allowing analysis and understanding of content captured by cameras in video sequences. Many applications can be cited, such as intelligent visual surveillance of human activities, intelligent visual surveillance of animal activities, optical motion capture, and multimedia applications. These diverse applications have generated a plethora of methods since the 1990's as they require to handle more and more complex challenges in more and more various environments.  In this context, the need for a framework to compare the performance of these algorithms is crucial. To this end, it is important to state correctly the problem.

For a given application of \emph{event detection}, one needs to define two aspects: what  the \emph{events} of interest are, and how  \emph{detection} is cast as a measurable task.
In videos, events can be defined at different spatial and temporal granularities.
For instance, at the \emph{pixel level}, background subtraction aims to detect pixels belonging to the foreground (event), as in the \CDNET benchmarks~\cite{Goyette2012Changedetection,Wang2014CDnet}.
At the \emph{frame level}, action spotting aims to detect the precise temporal anchor of an event in a video stream~\cite{Giancola2018SoccerNet}.
At the \emph{clip level}, applications operate over short video segments to detect event clips, as in fire and smoke detection~\cite{Hashemzadeh2022Smoke}, gas leak detection in infrared video~\cite{Wang2022VideoGasNet}, human intrusion detection~\cite{Lohani2022Perimeter}, or illegal waste dumping detection~\cite{Bouwmans2026Illegal}. 
In some cases, spatial and temporal bounds are difficult to specify, \eg, to delineate precisely a moving object, to spot fast-paced actions in sports, or, in contrast, a slowly evolving event may be interpreted as one event or several, depending on the use case~\cite{Lohani2022Perimeter}.
As a result, event detection can be cast as a two-class crisp classification task in which every prediction (on a pixel, on a frame, on a clip) independently yields one of four mutually exclusive outcomes: \emph{true negative}~($\outTN$, background ---\aka absence of event--- correctly not detected as an event), \emph{false positive}~($\outFP$, background incorrectly detected as an event), \emph{false negative}~($\outFN$, missed event), or \emph{true positive}~($\outTP$, event correctly detected).

Since part of the evaluation is based on probabilistic models (see \cref{sec:pipeline}), it cannot be stressed enough that the four outcomes must be carefully defined for each application. These considerations echo the paradox first highlighted by \citet{Bertrand1889Calcul}, namely that the probability distribution of events is closely tied to the precise definition of those events. Ambiguities easily arise, and intuition alone can lead to ill-posed formulations of an event detection task. 
For example, counting people in a video stream raises the same paradox. The question ``how many people entered the scene?'' admits at least two answers computed from the very same video, depending on the elementary event adopted. If the event is ``a person enters the field of view'', every entrance counts and a person who leaves and re-enters is counted twice. However, if the event is ``a \textit{new} person enters'', an entrance counts only when the identity has never been observed, which requires solving an open-set re-identification problem, the rejection case being known as novelty detection~\cite{BedagkarGala2014ASurvey}. As in Bertrand's paradox, the discrepancy would not originate in the measurement but in the under-specification of the event, and the question is simply ill-posed until the event is defined.

  This article focuses on establishing a rigorous framework for defining event detection tasks for videos. The main contributions are as follows:
 
\begin{enumerate}
    \item We present an original way of organizing data by hierarchically classifying videos according to the acquisition process and using tags to characterize the content. This allows for greater flexibility in selecting videos when developing or evaluating a method. This process is thoroughly explained in  \Cref{sec:dataset}, A.
    \item  In  \Cref{sec:dataset}, C, we describe  two new additional datasets: FSD, a  real-world dataset collected
from public IP cameras, and \datasetAnaisAcronym, a synthetic dataset with video clips acquired by fixed RGB surveillance cameras monitoring an urban crossroad.
    \item \Cref{sec:performance} presents a comprehensive evaluation protocol for a detection task. This protocol includes a pipeline ---disruptive in its design--- for evaluating a method’s probabilistic performance, as well as metrics related to hardware aspects.
   \item To facilitate the development of a method and an objective comparison, we propose using the concept of an application scenario. The elements of such a scenario are described in \cref{sec:scenarios}.
\end{enumerate}

%% file: sec/2_related_work_new.tex
\section{Related work}\label{sec:related_work}

From the very early days of event detection in videos,  whether it is considered at pixel-level, frame-level, or clip-level, the comparison of algorithms  has been achieved via a conventional implicit framework based on metrics and datasets. The metrics are based on ground-truth images using true positive, true negative, false positive, and false negative, providing precision, accuracy, and F-measure. 
As for datasets, enormous progress has been made with the evolution from small datasets to much larger ones. One of the most representative cases is the case of moving object detection using background subtraction. Indeed, datasets have been released for the evaluation of background subtraction since 1999 and have been developed from small-scale datasets up to large-scale datasets according to the need for fair comparison over large data covering diverse scenarios. In 1999, \citet{Toyama1999Wallflower} proposed the Wallflower dataset in the context of visual surveillance of human activities. 
This dataset, consisting of seven video sequences, with each sequence presenting one of the
difficulties a practical task is likely to encounter, was one of the most used
dataset in the period 2000-2010. Its main drawback is that there is only one ground-truth image per sequence. During the aforementioned period, several other datasets have been released, such as the I2R dataset, the LIMU dataset\footnote{\url{https://limu.ait.kyushu-u.ac.jp/dataset/en/}} and the VSSN 2006 dataset. 
Even if these datasets provide more ground-truth images than the Wallflower dataset, they  are still  limited in size. 
This absence of a single realistic large-scale dataset with accurate ground truth and providing a balanced coverage of the range of real-world challenges strongly affected the progress of research during at least one decade. 
No wonder then that since 2011 several  large datasets were created; most of them owe their appearance to the workshops where they were developed as part of the challenge and thus possess an associated evaluation measure. Between the most remarkable ones, we can cite (in chronological order):  
 \begin{itemize}
 \item The SABS dataset \cite{Brutzer2011Evaluation}, the ChangeDetection.net\footnote{\url{http://jacarini.dinf.usherbrooke.ca/}} (CDnet 2012, CDnet 2014), and BMC 2012\footnote{\url{https://backgroundmodelschallenge.eu/}} for foreground detection in color space. 
 \item SBM-RGBD dataset\footnote{\url{https://rgbd2017.na.icar.cnr.it/SBM-RGBDdataset.html}} for foreground detection in RGB-D space.
 \item SBI dataset\footnote{\url{https://sbmi2015.na.icar.cnr.it/}} and SBMCnet dataset\footnote{\url{http://scenebackgroundmodeling.net/}} for background initialization. 
\end{itemize}
The Stuttgart Artificial Background Subtraction (SABS) dataset, created for pixel-wise evaluation of the performance of background models for background subtraction, covers 9 typical challenges of background subtraction that occur in the context of video surveillance. The use of artificially generated data implies high-quality segmentation and ground-truth annotation  but naturally results in a lack of real-scenes. In 2012, the CDnet 2012 dataset  was developed as part of the CVPR 2012 Change Detection Workshop challenge (CDW 2012). It consists of 31 camera-captured videos (70,000 frames) spanning 6 categories selected to include diverse change and motion detection challenges, namely baseline, dynamic backgrounds, camera jitter, intermittent object motion, shadows and thermal category. In 2014, within the CVPR 2014 Change Detection Workshop challenge (CDW 2014), it was enlarged to form CDNet 2014. The 22 additional camera-captured videos (70,000 new frames) span 5 new categories that incorporate challenges that were not addressed in the 2012 dataset: Challenging Weather, Low Frame-Rate, night, PTZ, and air turbulence. In 2012, \citet{Vacavant2012ABenchmark} developed a benchmark dataset and an evaluation process built from both synthetic and real videos. This framework was used in the BMC workshop (Background Models Challenge) in conjunction with ACCV 2012. It focuses on outdoor situations with weather variations such as wind, sun, or rain. In addition, some evaluation criteria and an associated free software are provided to compute them from several challenging testing videos.

Although the aforementioned pairs dataset/performance evaluation have played a crucial role in advancing the field, they all suffer from the following limitations:
\begin{enumerate}
\item The datasets are limited to intelligent visual surveillance of human activities, mostly in urban scenes. In particular, they do not cover the case of visual surveillance of animal activities in natural scenes.
\item Their ranking methods are elementary, using a simple averaging method.
\item Results of both unsupervised and supervised methods are presented in the same ranking.
\end{enumerate}
The first limitation has been partially addressed by airport datasets (AGVS dataset\footnote{\url{http://www.agvs-caac.com/}}), maritime datasets (MAR dataset\footnote{\url{http://labrococo.diag.uniroma1.it/MAR/}}), and underwater datasets
(Fish4Knowledge\footnote{\url{https://groups.inf.ed.ac.uk/vision/DATASETS/FISH4KNOWLEDGE/}}), as well as several other datasets providing sequences captured with camera traps in natural environments (Caltech Camera Traps\footnote{\url{https://beerys.github.io/CaltechCameraTraps/}}, eMammal\footnote{\url{https://emammal.si.edu/}}). However, all these datasets are limited to their targeted applications, and they are not large enough for the training of deep learning methods. Expanding the range of datasets to be used is the main motivation for our study in \cref{sec:dataset}.

The second limitation has been addressed in the works of \citet{Pierard2025Foundations,Pierard2024TheTile-arxiv} which have been applied successfully in the International Contest on Illegal Waste Dumping Detection (IWSS 2026) in conjunction with WACV 2026~\cite{Bouwmans2026Illegal}. These works have led us to propose the performance evaluation tools described in \cref{sec:performance}.

The last limitation, the mixing of results for unsupervised and supervised methods in the same ranking, is more complex. It requires to define precisely the meaning of unsupervised vs. supervised setups (one could argue that as soon as ground truths are released, a method automatically switches to supervised-like mode). This only serves to further clarify the terms of an application scenario, which is the subject of \cref{sec:scenarios}.

%% file: sec/3_dataset.tex
\section{Dataset description}\label{sec:dataset}
The purpose of this section is twofold: 1) describe the content of existing datasets, 2) present a way to organize raw data, annotations, and metadata in such a way that each video can be used in different scenarios.

\subsection{Event-Monitoring Video Dataset (v1): A Novel Structuration Methodology}
The proposed large-scale event-monitoring video dataset (Version 1) can be categorized according to the environments in which the data are captured. In this context, the full dataset includes four main real-life scenarios: urban, natural/wildlife, maritime, and underwater environments to cover a long spectrum of visual applications in diverse environments. The challenges associated with these videos vary depending on both the environment and the acquisition setup. For example, some videos are captured using CCTV cameras, whilst others may include thermal data. Therefore, the difficulty of event monitoring is influenced not only by the scene content but also by the sensing modality and camera configuration. For this reason, we organize the videos according to the environment in which they are captured. We innovate about the dataset annotation and the structuration of dataset annotation as explained in Section \ref{PubliclyAvailableDatasets} and in Section \ref{AdditionalPublic/PrivateDatasets}.

The novelty of the proposed dataset lies in its unified structuration of event-monitoring videos across heterogeneous environments, rather than in focusing on a single scene type, modality, or acquisition condition. Existing benchmarks are often designed for specific domains, acquisition setups, or isolated visual challenges. Moreover, they frequently follow different organizational formats, annotation conventions, metadata structures, and category definitions. This heterogeneity makes it difficult to integrate them into a common learning framework and limits the ability of deep learning models to learn consistent representations across diverse environments. To address this limitation, the proposed dataset brings together urban, natural/wildlife, maritime, and underwater scenarios within a common organizational framework. Each video is structured according to its environment, sensing modality, and camera configuration. In addition, the visual and acquisition-related difficulties present in each sequence are explicitly represented as metadata tags. This tag-based annotation strategy enables videos to be selected, grouped, and evaluated according to specific challenges, such as illumination variation, dynamic backgrounds, occlusion, camera motion, low contrast, camouflage, sensor fusion, and marine effects. This structuration supports both broad cross-environment evaluation and more targeted challenge-specific analysis. It allows deep learning models to be trained and assessed under diverse visual conditions, thereby encouraging the learning of more robust and transferable object representations across domains. At the same time, the metadata tags provide a systematic way to analyze model performance with respect to the specific difficulties present in each sequence.

\begin{figure*}[t]
\centering
\resizebox{\textwidth}{!}{%
\begin{tikzpicture}[
    box/.style={
        draw,
        rectangle,
        rounded corners,
        align=center,
        font=\small,
        inner sep=4pt
    },
    edge/.style={draw}
]

\node[box] (root) at (0,0) {Event Monitoring for Video Datasets (EMVD)};

\node[box] (urban) at (-10,-1.6) {Urban};
\node[box] (natural) at (-3.2,-1.6) {Natural};
\node[box] (underwater) at (3.2,-1.6) {Underwater};
\node[box] (maritime) at (10,-1.6) {Maritime};

\draw[edge] (root) -- (urban);
\draw[edge] (root) -- (natural);
\draw[edge] (root) -- (underwater);
\draw[edge] (root) -- (maritime);

\node[box] (urban-rgb) at (-13.2,-3.2) {RGB};
\node[box] (urban-ir) at (-10.7,-3.2) {Thermal / IR};
\node[box] (urban-rgbd) at (-8.1,-3.2) {RGB-D};
\node[box] (urban-syn) at (-5.7,-3.2) {Synthetic};

\draw[edge] (urban) -- (urban-rgb);
\draw[edge] (urban) -- (urban-ir);
\draw[edge] (urban) -- (urban-rgbd);
\draw[edge] (urban) -- (urban-syn);

\node[box] (urban-moving) at (-14.6,-4.9) {Moving camera};
\node[box] (urban-stationary) at (-11.8,-4.9) {Stationary camera};

\draw[edge] (urban-rgb) -- (urban-moving);
\draw[edge] (urban-rgb) -- (urban-stationary);

\node[box] (natural-rgbir) at (-3.2,-3.2) {RGB + IR};

\draw[edge] (natural) -- (natural-rgbir);

\node[box] (underwater-rgb) at (3.2,-3.2) {RGB};

\draw[edge] (underwater) -- (underwater-rgb);

\node[box] (maritime-rgb) at (6.4,-3.2) {RGB};
\node[box] (maritime-ir) at (8.9,-3.2) {Thermal / IR};
\node[box] (maritime-multi) at (11.6,-3.2) {Multi-modal};
\node[box] (maritime-syn) at (14.2,-3.2) {Synthetic};

\draw[edge] (maritime) -- (maritime-rgb);
\draw[edge] (maritime) -- (maritime-ir);
\draw[edge] (maritime) -- (maritime-multi);
\draw[edge] (maritime) -- (maritime-syn);

\node[box] (maritime-moving) at (5.0,-4.9) {Moving camera};
\node[box] (maritime-stationary) at (7.8,-4.9) {Stationary camera};

\draw[edge] (maritime-rgb) -- (maritime-moving);
\draw[edge] (maritime-rgb) -- (maritime-stationary);

\end{tikzpicture}%
}
\caption{Categorization of the Event-Monitoring Video Dataset According to Environment and Modality}
\label{fig:emvd_dataset_taxonomy}
\end{figure*}
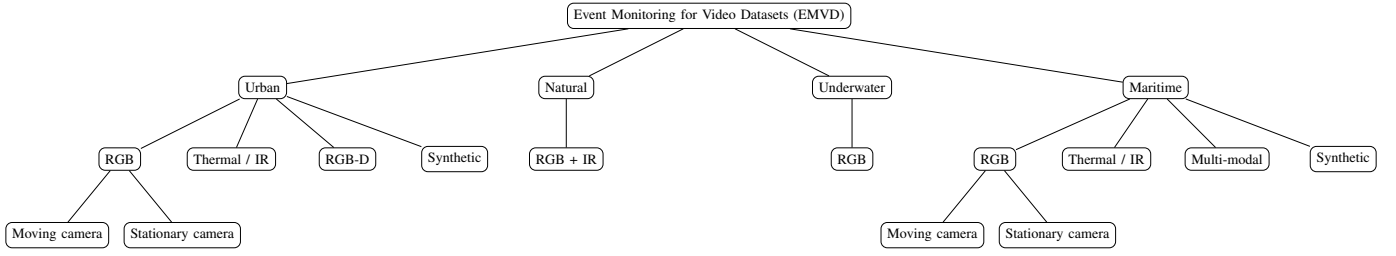

\begin{table*}[t]
\centering
\caption{Challenges associated with Different Modalities and Environments.}
\label{tab:environment_modality_dataset}
\scriptsize
\setlength{\tabcolsep}{3pt}
\renewcommand{\arraystretch}{1.18}

\begin{tabularx}{\textwidth}{
p{1.6cm}
p{1.5cm}
p{2.1cm}
>{\raggedright\arraybackslash}X
>{\raggedright\arraybackslash}X
}
\toprule
\textbf{Environment} & \textbf{Modality} & \textbf{Camera type} & \textbf{Tags} & \textbf{Datasets} \\
\midrule

\multirow{5}{*}{Urban}
& \multirow{2}{*}{RGB}
& Stationary camera
& Dynamic background, shadows, intermittent motion, illumination change, occlusion
& \CDNET 2014, PETS, i-LIDS, UCSD, ETISEO, BMC, ATON, SZTAKI, SBI, SBMnet, GTFD, AGVS, \fsdAcronym \\

& & Moving camera
& Camera jitter, scale variation, PTZ camera
& \CDNET 2014, UCSD, SBI, SBMnet, AGVS, VIRAT, ATON, Audio-Visual Vehicle \\

& Thermal / IR
& Stationary thermal camera
& Low contrast, video noise
& OSU Thermal, Terravic Motion IR, GTFD, Remote Scene IR \\

& RGB-D
& Stationary camera
& Depth camouflage, inserted background
& CITIC RGB-D, SBM-RGBD \\

& Synthetic
& Stationary camera
& Dynamic background, shadows, camera motion
& SABS, SYNTHIA, MOTSynth, SUC \\

\midrule

Natural
& RGB + IR
& Stationary camera
& Illumination variation, camouflage
& Caltech Camera Traps, eMammal, CAMO-UOW \\

\midrule

Underwater
& RGB
& Stationary camera
& Dynamic background, complex background, crowded scenes, marine effects
& Aqu@theque, Fish4Knowledge, Realworld Underwater Image Enhancement (RUIE) \\

\midrule

\multirow{5}{*}{Maritime}
& \multirow{2}{*}{RGB}
& Stationary camera
& Dynamic water background, illumination change, scale variation
& Ships Dataset, MASATI, Singapore Maritime Dataset (SMD) \\

& & Moving camera
& Dynamic water background, scale variation
& Maritime Obstacle Detection Dataset (MODD), Singapore Maritime Dataset (SMD) \\

& Thermal / IR
& Stationary thermal camera
& Low thermal contrast, thermal noise
& MassMIND \\

& Multimodal
& Moving camera
& Sensor fusion, small object detection
& M3FD\_Fusion, MarDCT \\

& Synthetic
& Stationary camera
& Controlled sea states, sea clutter
& Maritime Synthetic Dataset \\

\bottomrule
\end{tabularx}
\end{table*}

\subsection{Publicly Available Datasets}
\label{PubliclyAvailableDatasets}
\begin{figure*}
\begin{center}
\includegraphics[width=\textwidth]{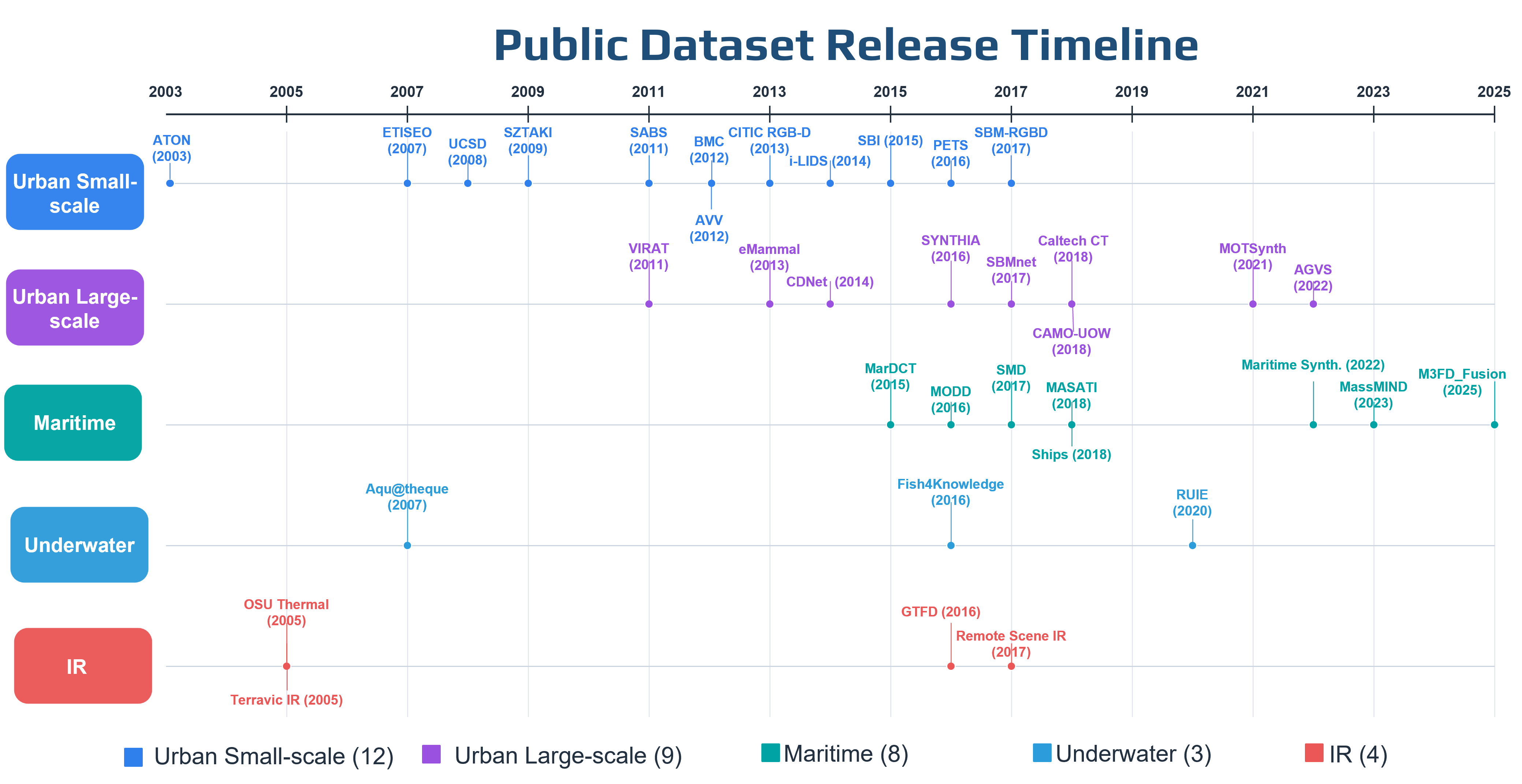}
\caption{Event Detection in Videos Datasets Timeline for Each Category: Urban Small-scale datasets, Urban Large-scale datasets, Maritime datasets, Underwater datasets, and IR datasets.}
\label{Overview-Datasets}
\end{center}
\end{figure*}

We consider 36 publicly available datasets captured across different kind of environments and modalities: their historical timeline (by category) is given by \cref{Overview-Datasets}. The complete description of each dataset could be found in \cref{sec:appendix-data} of the appendix.
\begin{itemize}
\item \textbf {Content.} 
\begin{itemize}
\item \textbf{Urban environments:} ATON (2003), OSU Thermal (2005), Terravic Motion IR (2005), ETISEO (2007), UCSD (2008), SZTAKI (2009), VIRAT (2011), SABS (2011), BMC (2012), CITIC RGB-D (2013), \CDNET (2014), i-LIDS (2014), SBI (2015), PETS (2016), GTFD (2016), SYNTHIA (2016), SBMnet (2017), Remote Scene IR (2017), SBM-RGBD (2017), MOTSynth (2021), AGVS (2022), Audio-Visual Vehicle.
 
\item \textbf{Natural environments:} eMammal (2013), Caltech Camera Traps (2018), CAMO-UOW (2018).

\item \textbf{Maritime environments:} MarDCT (2015), Maritime Obstacle Detection Dataset (MODD) (2016), Singapore Maritime Dataset (SMD) (2017), Ships Dataset (2018), MASATI (2018), Maritime Synthetic Dataset (2022), MassMIND (2023), M3FD\_Fusion (2025).

\item \textbf{Underwater environments:} Aqu@theque (2007), Fish4Knowledge (2016), Realworld Underwater Image Enhancement (RUIE) (2020).

\end{itemize}
\item \textbf{Description of the annotations.}
The data is organized in the following environments:
\begin{itemize}
\item \textbf{Urban environments:} includes scenarios captured in areas with humans, highways, buildings, and related structures.
\item \textbf{Natural environments:}  includes scenarios in which animals are captured in their natural habitats.
\item \textbf{Maritime environments:} consists of videos involving boats and ships, with images captured using satellite imagery or standard imagery from the shore.
\item \textbf{Underwater environments:} Videos present additional challenges due to light degradation, color degradation, camouflage, and related factors, which make object detection more difficult.
\end{itemize}

In each scenario, the data are collected using different modalities, such as:
\begin{itemize}
\item \textbf{RGB data:} RGB data contains the videos captured by RGB camera. Additionally, the camera could be moving or static so it  is further divided into two categories: stationary-camera RGB videos and moving-camera RGB videos. These modalities include several challenges, such as dynamic backgrounds, camera jitter, shadows, illumination changes, occlusion, and scale variations.
\item \textbf{RGB-D data:} RGB-D data include depth information in addition to RGB videos. Depth camouflage and inserted backgrounds increase the complexity of detection in this modality.
\item \textbf{Infrared data:} The main challenges in the infrared modality are low contrast and video noise.
\item \textbf{Synthetic data:} Synthetic scenes may include various challenges similar to those found in RGB data.
\end{itemize}
\item \textbf{Associated tags.} The annotation structure depends on the scenario and will be presented together with the scenarios. \Cref{tab:environment_modality_dataset} summarizes the challenges associated with different modalities and environments. The camera configuration may be either stationary or moving, and this directly affects the nature of the challenges involved in object identification. 
In stationary-camera settings, the environment mainly induces the challenges. For instance, background motion caused by trees, grass, or similar elements can generate dynamic background patterns that must be distinguished from the target object. In addition, shadows and illumination changes may occur due to variations in the angle and intensity of the light source. Objects may also exhibit intermittent motion when they remain static for some time, or they may become partially or fully occluded by other scene elements. In moving-camera settings, additional challenges arise due to camera jitter, object scale variation, and pan-tilt-zoom camera motion. Some datasets are captured using thermal or infrared cameras, where low contrast and video noise are the main challenges. RGB-D data, which combine RGB information with depth information, introduce additional difficulties such as depth camouflage and inserted backgrounds. In maritime environments, multimodal data may also be used, where information from multiple sensors is combined. This introduces challenges related to sensor fusion and small object detection.
\end{itemize}

\subsection{Additional Public/Private Datasets}
\label{AdditionalPublic/PrivateDatasets}
Our study includes two datasets that we created recently and whose content is partly public: (1) the \fsd (\fsdAcronym) and (2) the \datasetAnaisName (\datasetAnaisAcronym) dataset; parts of these datasets are kept private to enable a fair evaluation.

\subsubsection{Description of the \fsd dataset} 
The \fsd (\fsdAcronym) is a real-world dataset collected from public IP cameras. It was designed to evaluate video understanding and foreground segmentation methods under diverse surveillance conditions. The dataset includes a wide range of indoor and outdoor scenes, with variations in illumination, object scale, scene activity, camera behavior, and recording duration. The main features of the dataset can be enumerated as follows.

\begin{itemize}
\item \textbf{Content of the FSD dataset.}
The dataset consists of video sequences acquired from $43$ public IP cameras. Each camera folder contains RGB frames and their corresponding ground-truth foreground masks. When available, a reference background image is also provided for the corresponding camera sequence. The dataset is organized at the camera level, where each camera directory contains a \texttt{frames} folder for RGB images, a \texttt{masks} folder for foreground annotations, and optionally a \texttt{background.jpg} image. Overall, \fsdAcronym contains $153{,}191$ annotated frame--mask pairs. The number of frames varies significantly across cameras, ranging from $23$ frames to $51{,}800$ frames, with an average of approximately $3{,}563$ frames per camera. All $43$ camera folders have matching numbers of RGB frames and foreground masks, which enables consistent quantitative evaluation at both frame and camera levels.

\input{figs/fsd_fig}

\item \textbf{Description of the annotations for the \fsdAcronym dataset.}
The annotations consist of pixel-level foreground masks associated with the RGB frames. These masks identify the foreground regions in each frame and can be used to evaluate foreground segmentation, motion segmentation, and related video analysis methods. The dataset therefore provides direct supervision for separating moving or relevant foreground objects from the background scene. In addition to the frame-level foreground masks, each camera is associated with one or more challenge tags. These tags describe the visual conditions, scene content, and acquisition properties of the corresponding camera sequence. They can be used to construct balanced training, validation, and test splits, to stress-test models under specific conditions, and to report performance according to different types of challenges.

\item \textbf{Tags associated to the \fsdAcronym dataset.}
The camera-level tags cover both scene content and visual challenges. The most common tags are \textit{cars}, which appear in $30$ cameras, and \textit{pedestrians}, which appear in $15$ cameras. Other frequent tags include \textit{illumination change} and \textit{long video}, each appearing in $10$ cameras, followed by \textit{few objects} in $9$ cameras and \textit{small objects} in $6$ cameras. The dataset also includes more specific challenges such as \textit{static objects}, \textit{many objects}, \textit{night vision}, \textit{very long video}, \textit{animals}, \textit{camera switch}, \textit{no objects}, and \textit{boats}. These tags are not mutually exclusive, since a single camera sequence may contain multiple challenges. This diversity makes \fsdAcronym suitable for evaluating the robustness of computer vision methods under realistic public-camera conditions, including changes in illumination, object size, object density, nighttime acquisition, and challenging foreground-background separation scenarios.
\end{itemize}

\subsubsection{Description of the \datasetAnaisName (\datasetAnaisAcronym) dataset}

\datasetAnaisName (\datasetAnaisAcronym) is a synthetic dataset generated using the CARLA simulator (version 0.9.15)~\cite{Dosovitskiy2017CARLA}. CARLA provides pixel-perfect instance segmentation annotations together with fine-grained control over environmental conditions, making it particularly suitable for the evaluation of computer vision algorithms under diverse urban scenarios. The main features of the dataset can be enumerated as follows.
\begin{itemize}
\item \textbf{Content of the \datasetAnaisAcronym dataset.} The dataset consists of video clips acquired by fixed RGB surveillance cameras monitoring an urban crossroad. Camera viewpoints vary across sequences, while the recorded scenes capture a wide range of illumination and weather conditions throughout a full day-night cycle (see \cref{fig:rgb_sequences}). In addition to RGB videos (encoded in \texttt{}{MPEG-4}  format), the dataset provides pixel-level instance segmentation ground-truth masks (\texttt{MKV} format) and metadata describing the weather conditions for each frame in each sequence (\texttt{JSON} format). \Cref{fig:rgb_vs_instance} illustrates an RGB frame with its corresponding instance mask. Furthermore, the dataset includes a Python script for automatically generating binary motion segmentation masks from the instance segmentation annotations, enabling the separation of foreground and background regions.

\input{figs/rgb_vs_instance}

Six different cameras were used, alternating, to film the scene. All data were recorded at a frame rate of $25$ frames per second with a HD/720p spatial resolution of $1280 \times 720$ pixels. The dataset contains $25$ video sequences of $5$ minutes each, corresponding to a total duration of $125$ minutes. Each sequence comprises $7{,}500$ frames, resulting in a total of $187{,}500$ annotated frames.

\item \textbf{Description of the annotations for the \datasetAnaisAcronym dataset.} 
The semantic masks were annotated as follows. 
CARLA provides $29$ semantic classes, including: Road, Sidewalk, Building, Wall, Fence, Pole, Traffic Light, Traffic Sign, Vegetation, Terrain, Sky, Pedestrian, Rider, Car, Truck, Bus, Train, Motorcycle, Bicycle, Water, Road Line, Bridge, Rail Track, and Guard Rail, among others. %
In the instance segmentation masks, the semantic class is encoded in the red channel, while the combination of the green and blue channels uniquely identifies each object instance. In addition to the instance segmentation annotations, binary motion segmentation masks can be automatically generated to distinguish foreground from background regions. To this end, we provide a Python script that annotates pixels of objects belonging to dynamic classes (\eg, pedestrians, riders, cars, trucks, motorcycles, and bicycles) as the foreground, whereas pixels of static scene elements (\eg, roads, sidewalks, buildings, vegetation, traffic signs, and traffic lights) are assigned to the background.  This annotation assumes that no objects of the dynamic class labels remain static for the duration of the video clip, which we enforced during the simulation process. 

\item \textbf{Tags associated to the \datasetAnaisAcronym dataset.} 
The sequences depict realistic urban traffic scenes containing pedestrians walking on sidewalks and crossing roads, as well as vehicles such as cars, trucks, motorcycles, and bicycles moving through the intersection. The dataset includes challenging visual phenomena commonly encountered in real-world surveillance applications, such as cast shadows, moving vegetation, reflections on wet road surfaces, and water accumulation during heavy rainfall.

Weather conditions include clear, rainy, foggy, and windy weather, as well as combinations of these conditions. The $25$ sequences span a complete day-night illumination cycle, covering daylight, dusk, dawn, and nighttime scenarios. This diversity makes the dataset suitable for evaluating the robustness of computer vision methods to changes in weather, illumination, viewpoint, and scene dynamics.
\end{itemize}

\input{figs/rgb_sequences}

%% file: figs/fsd_fig.tex
\begin{figure}[th]
    \centering
    \begin{minipage}{0.98\linewidth}
        \centering
        \includegraphics[width=\linewidth]{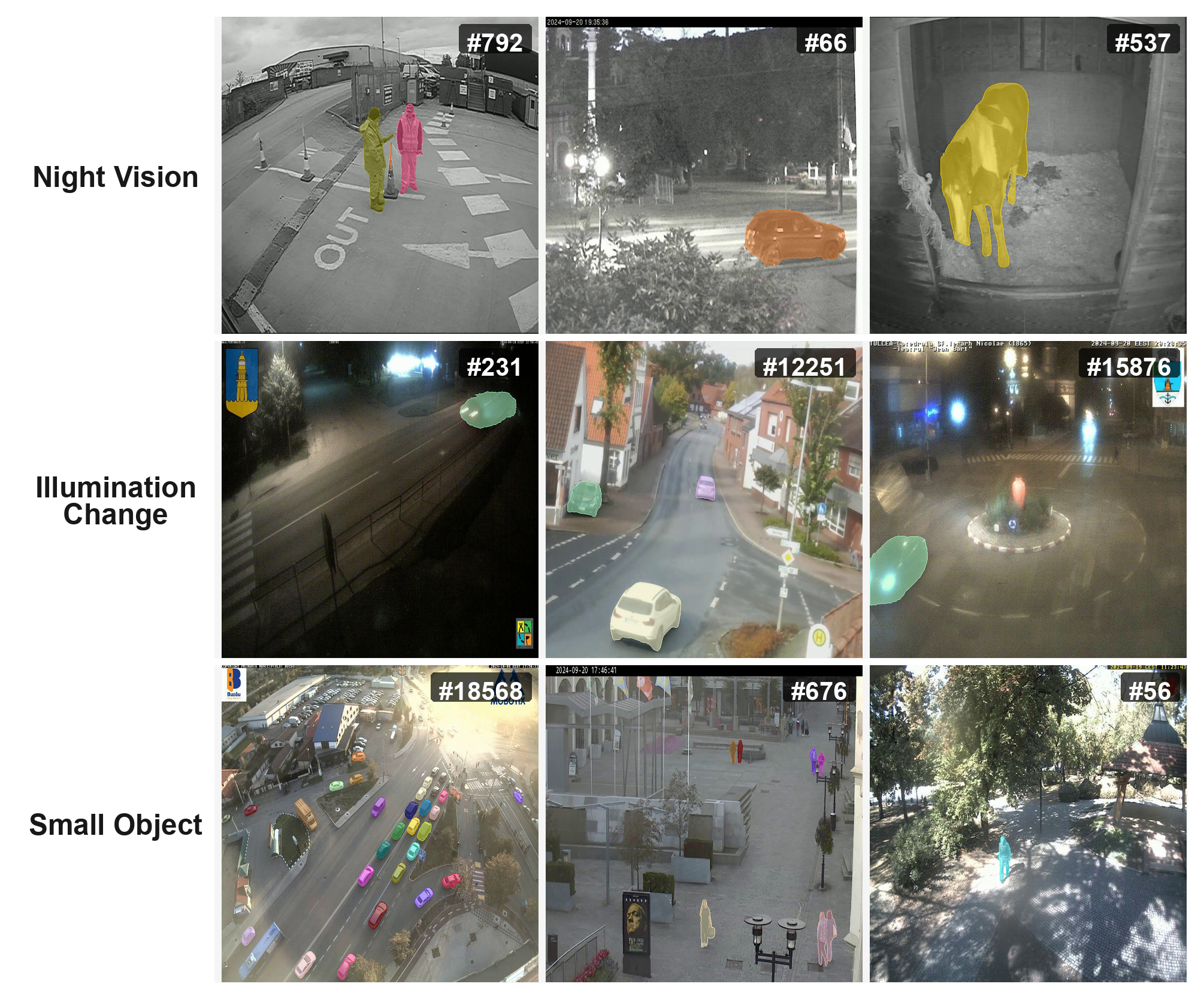}
    \end{minipage}

    \caption{Representative samples from the \fsd (\fsdAcronym) under challenging public IP camera conditions. Each of three rows corresponds to one of the challenges: night vision, illumination change, and small objects. Each column shows a different RGB frame with the ground-truth foreground segmentation masks overlaid; different colors indicate separate foreground objects, and frame numbers appear in the top-right corner.}
    \label{fig:fsd}
\end{figure}

%% file: figs/rgb_vs_instance.tex
\begin{figure}[th]
    \centering
    \begin{minipage}{0.49\linewidth}
        \centering
        \includegraphics[width=\linewidth]{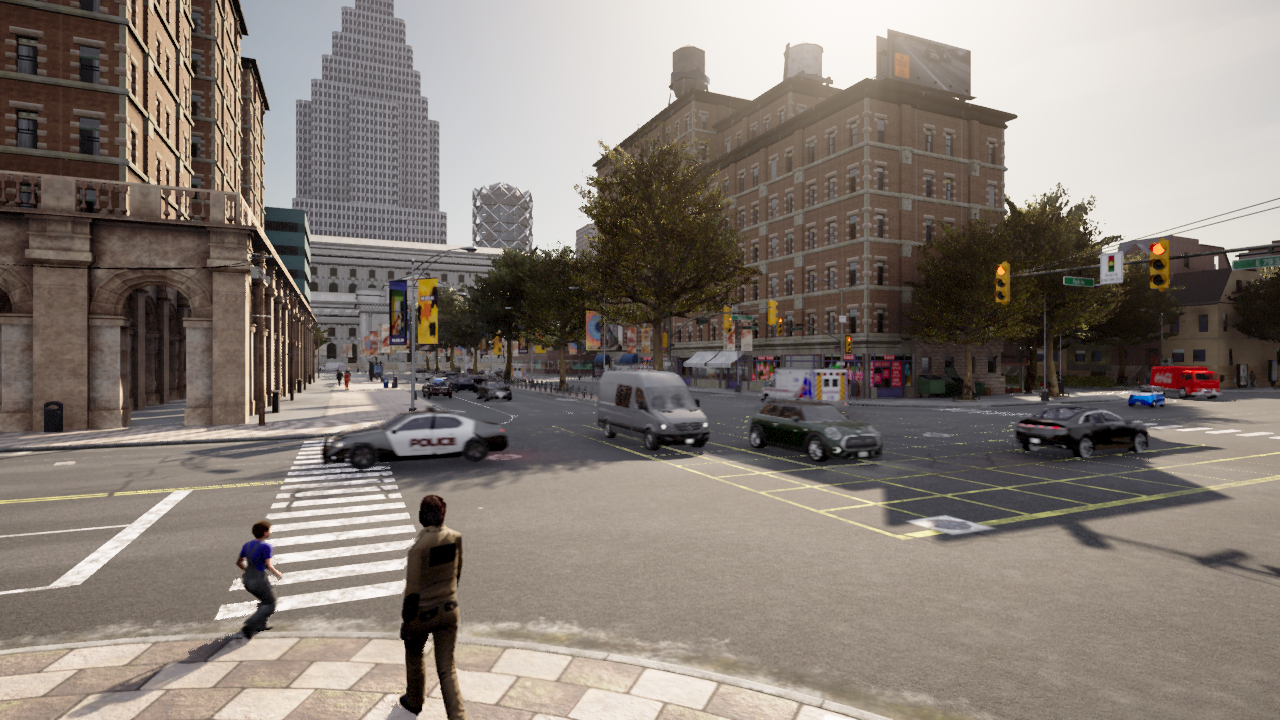}
    \end{minipage}
    \hfill
    \begin{minipage}{0.49\linewidth}
        \centering
        \includegraphics[width=\linewidth]{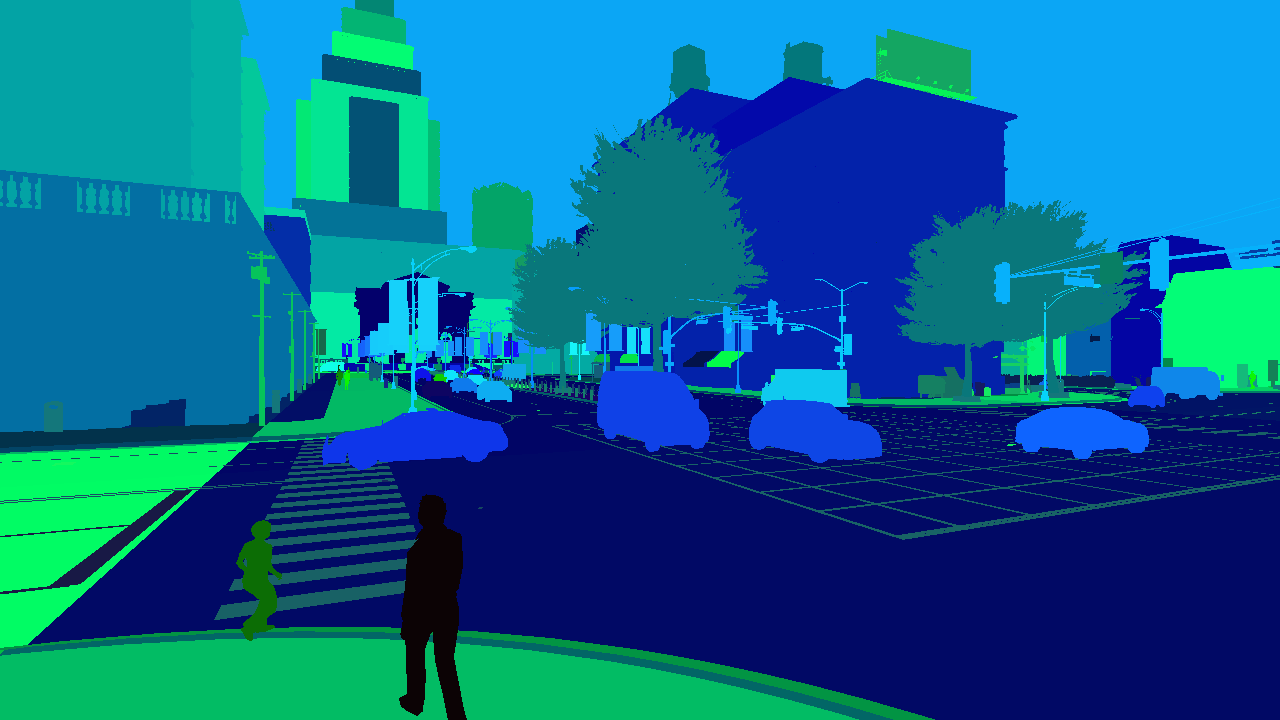}
    \end{minipage}

    \caption{An RGB frame with its corresponding instance mask (from the \datasetAnaisAcronym dataset).}
    \label{fig:rgb_vs_instance}
\end{figure}

%% file: figs/rgb_sequences.tex
\begin{figure*}[!tbp]
    \centering
    \begin{minipage}{0.19\linewidth}
        \centering
        \includegraphics[width=\linewidth]{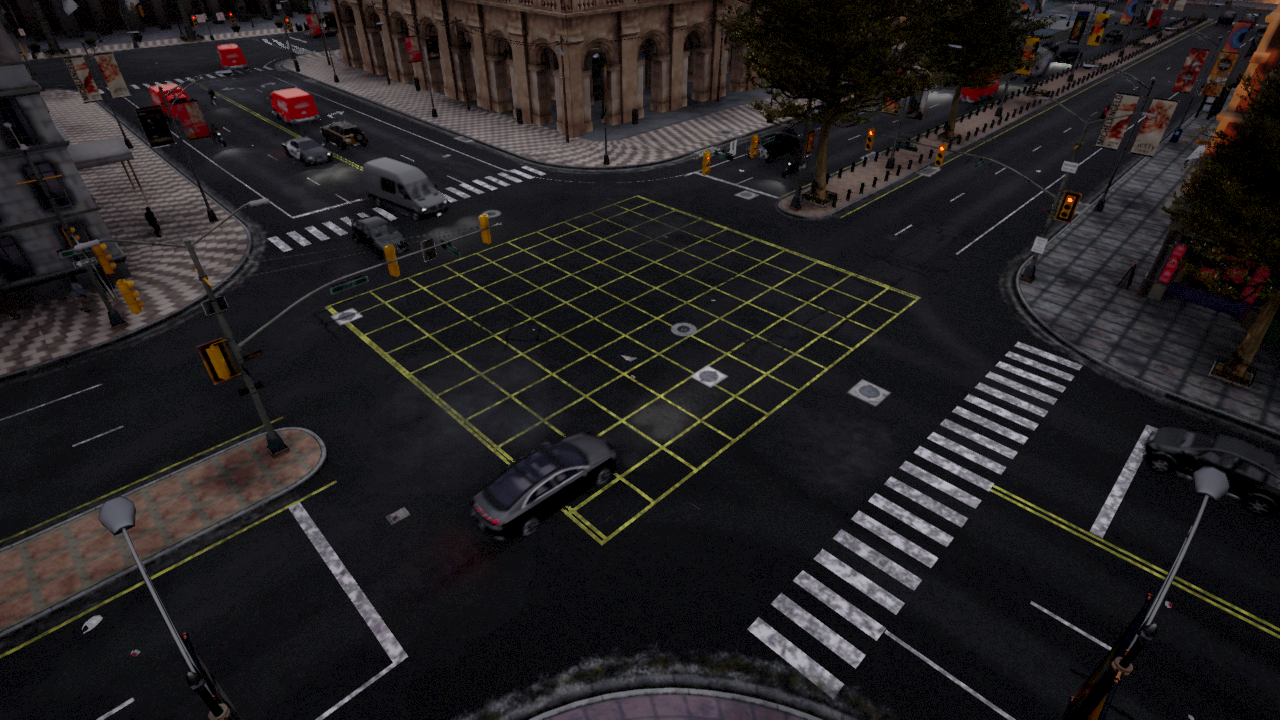}
    \end{minipage}
    \hfill
    \begin{minipage}{0.19\linewidth}
        \centering
        \includegraphics[width=\linewidth]{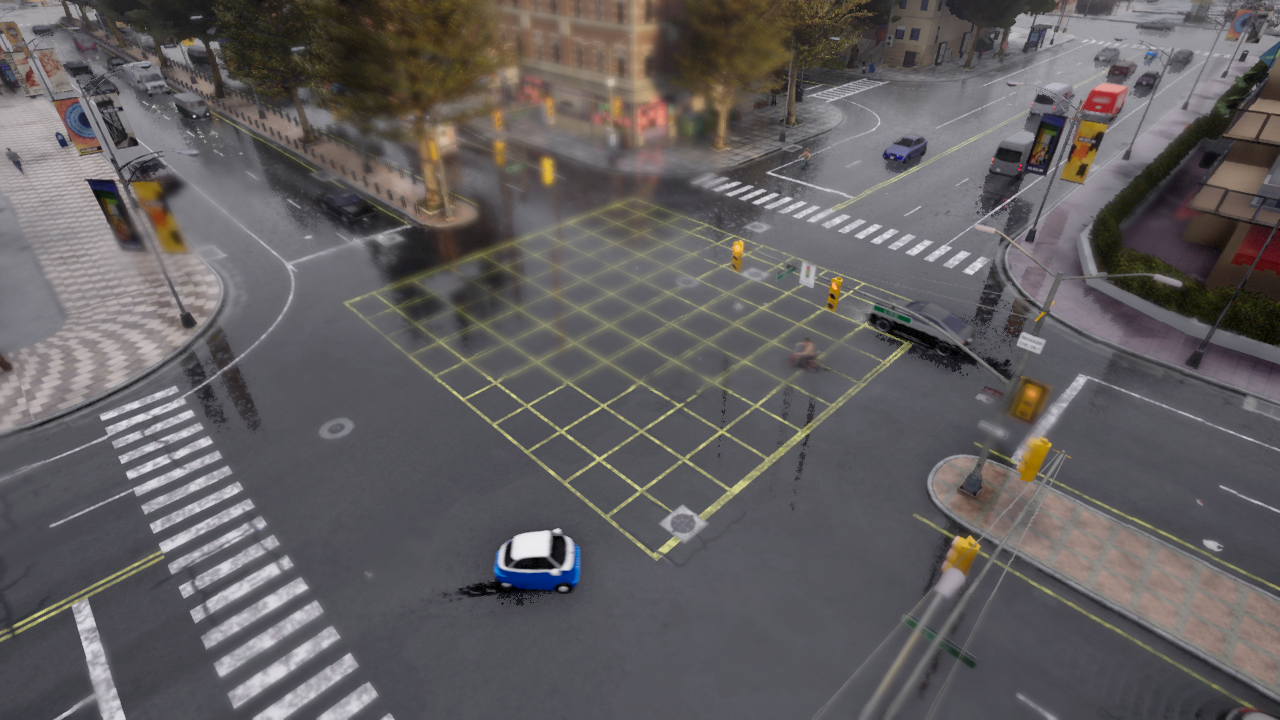}
    \end{minipage}
    \hfill
    \begin{minipage}{0.19\linewidth}
        \centering
        \includegraphics[width=\linewidth]{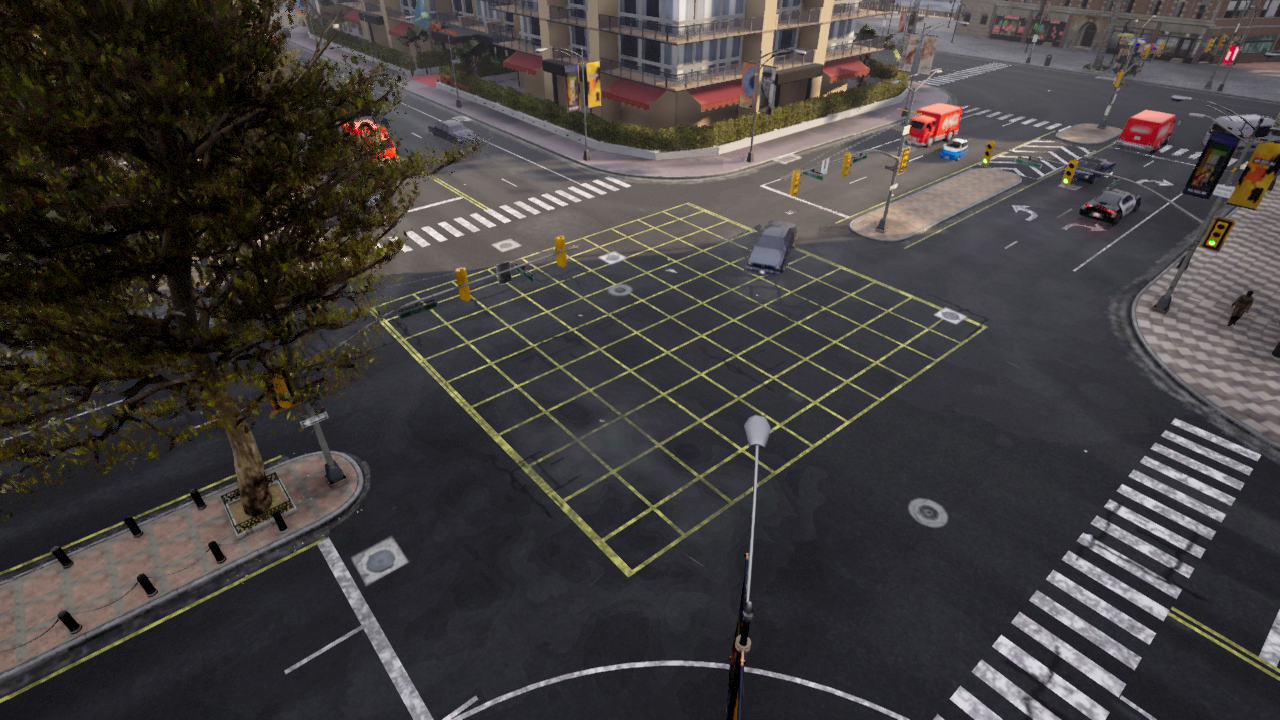}
    \end{minipage}
    \hfill
    \begin{minipage}{0.19\linewidth}
        \centering
        \includegraphics[width=\linewidth]{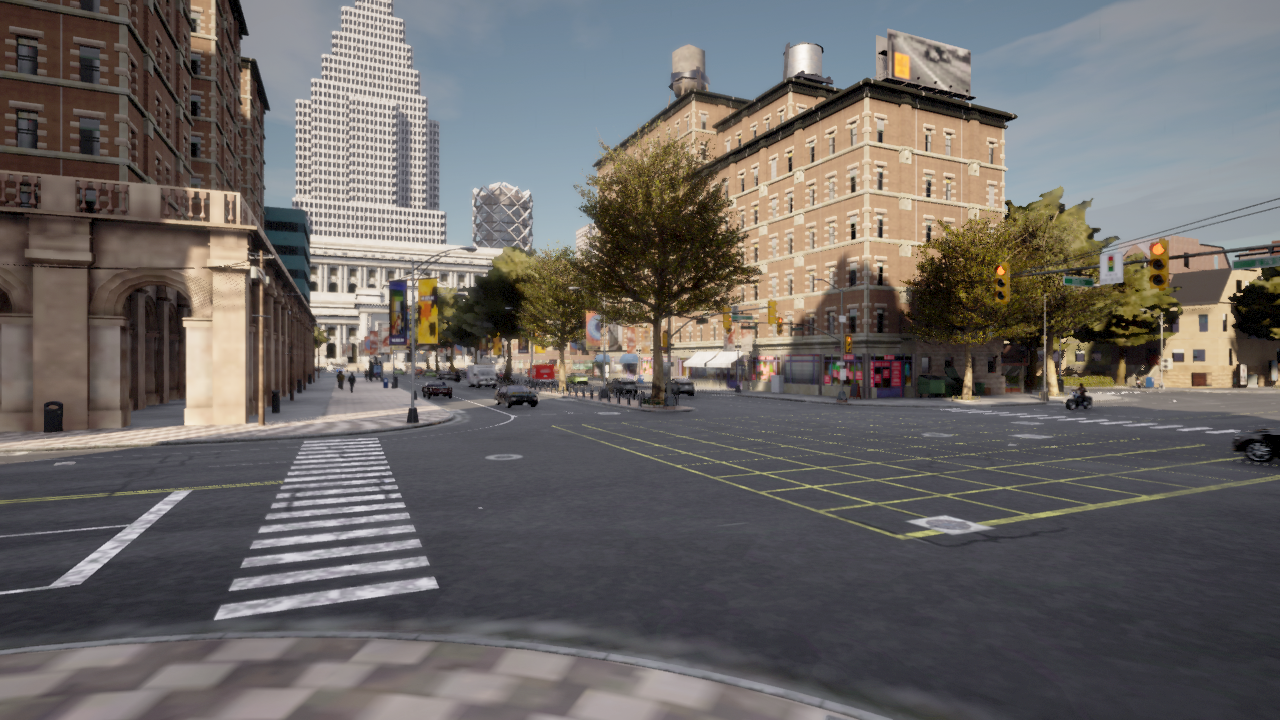}
    \end{minipage}
    \hfill
    \begin{minipage}{0.19\linewidth}
        \centering
        \includegraphics[width=\linewidth]{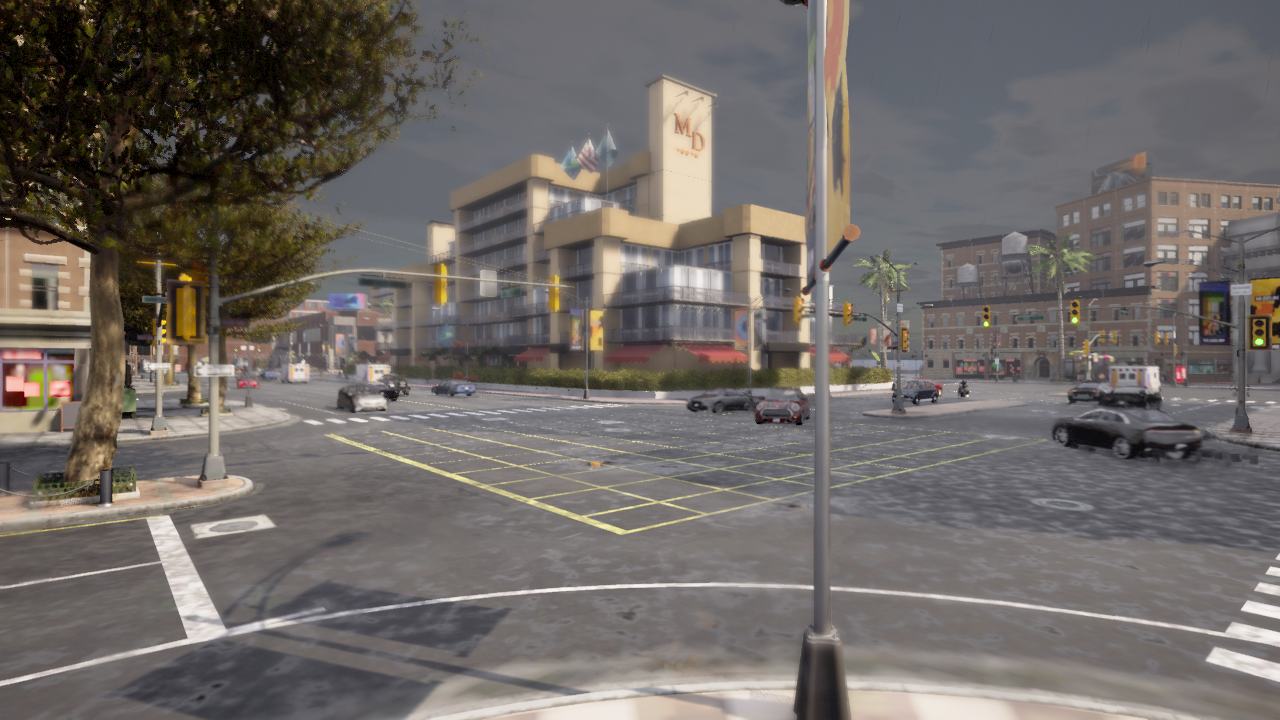}
    \end{minipage}

    \vspace{.5em}
    
    \begin{minipage}{0.19\linewidth}
        \centering
        \includegraphics[width=\linewidth]{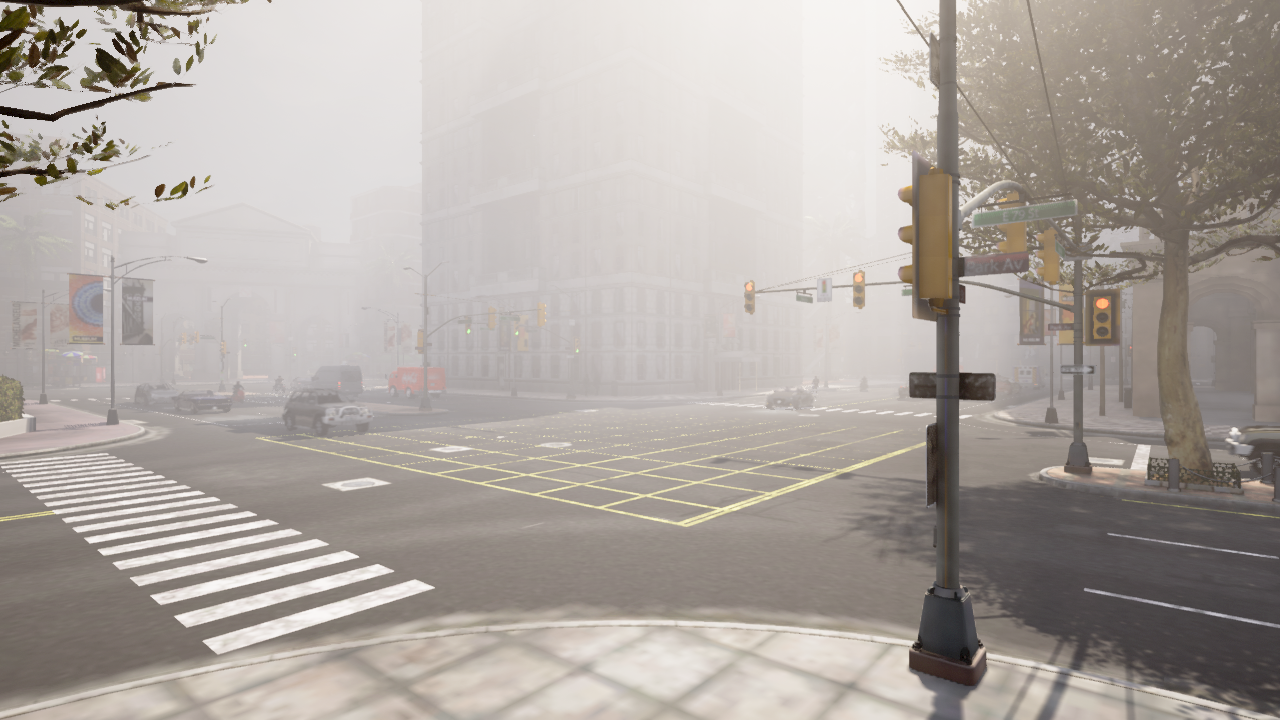}
    \end{minipage}
    \hfill
    \begin{minipage}{0.19\linewidth}
        \centering
        \includegraphics[width=\linewidth]{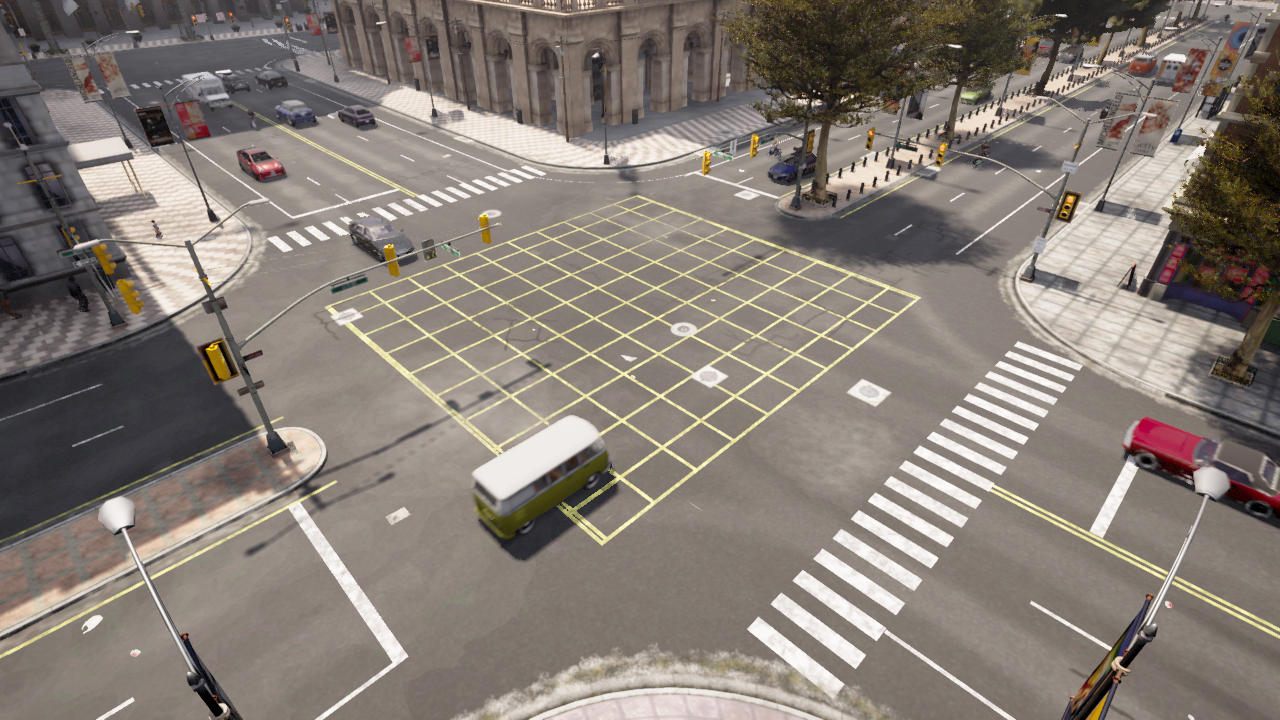}
    \end{minipage}
    \hfill
    \begin{minipage}{0.19\linewidth}
        \centering
        \includegraphics[width=\linewidth]{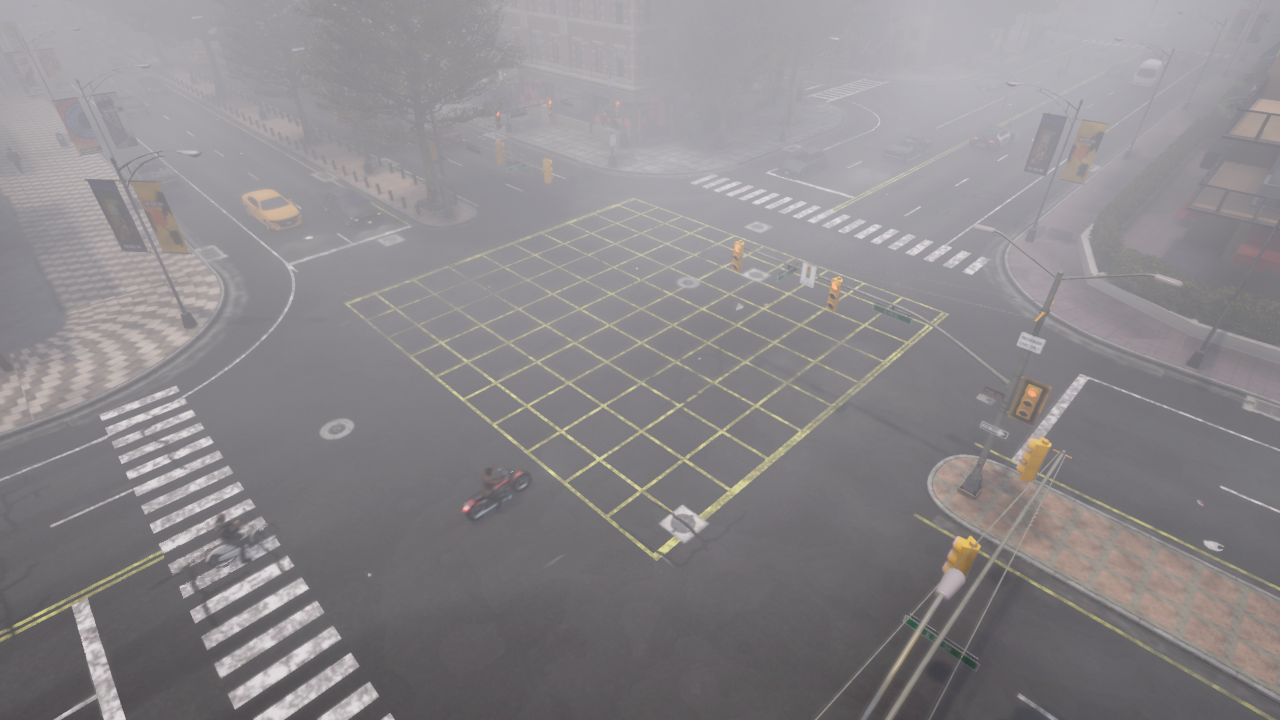}
    \end{minipage}
    \hfill
    \begin{minipage}{0.19\linewidth}
        \centering
        \includegraphics[width=\linewidth]{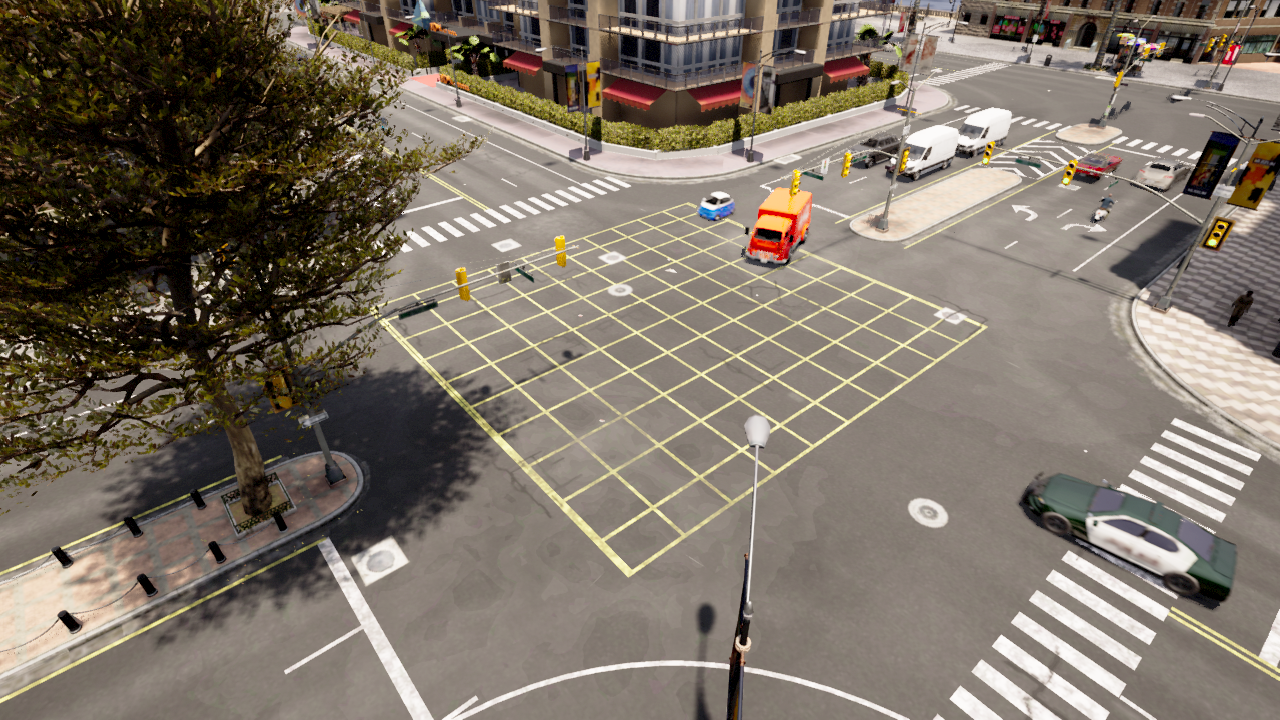}
    \end{minipage}
    \hfill
    \begin{minipage}{0.19\linewidth}
        \centering
        \includegraphics[width=\linewidth]{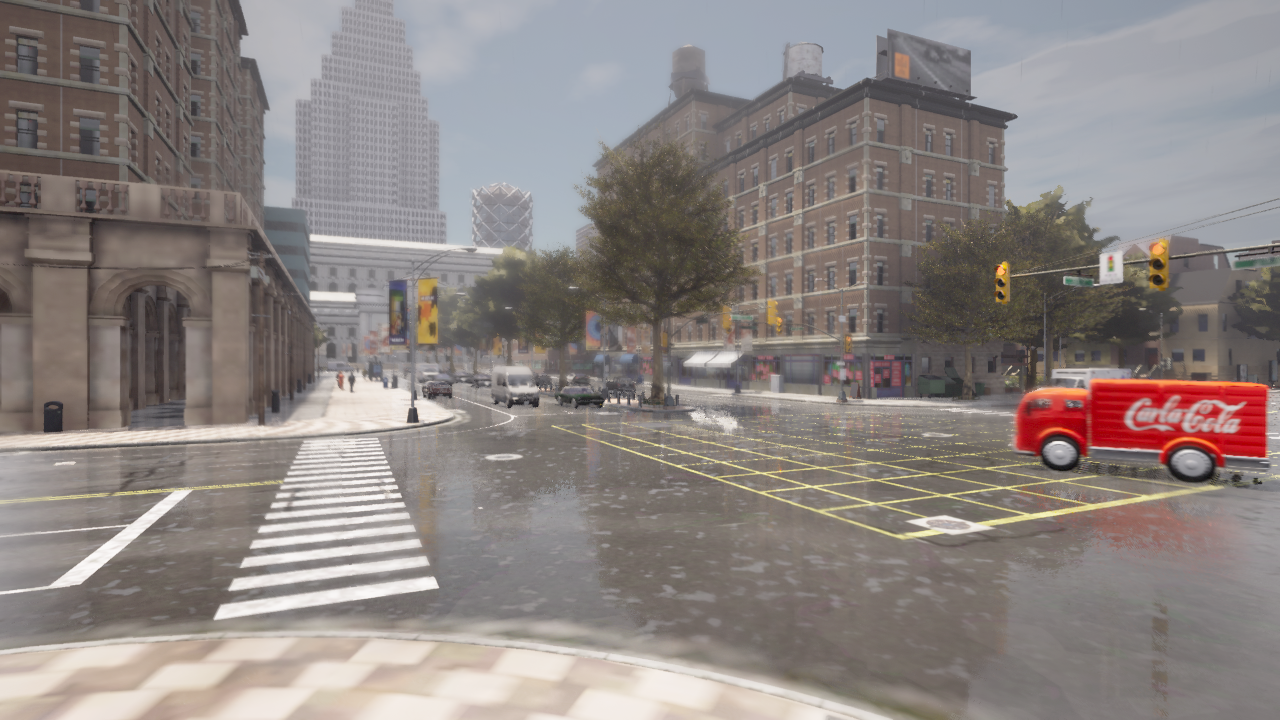}
    \end{minipage}

    \vspace{.5em}
    
    \begin{minipage}{0.19\linewidth}
        \centering
        \includegraphics[width=\linewidth]{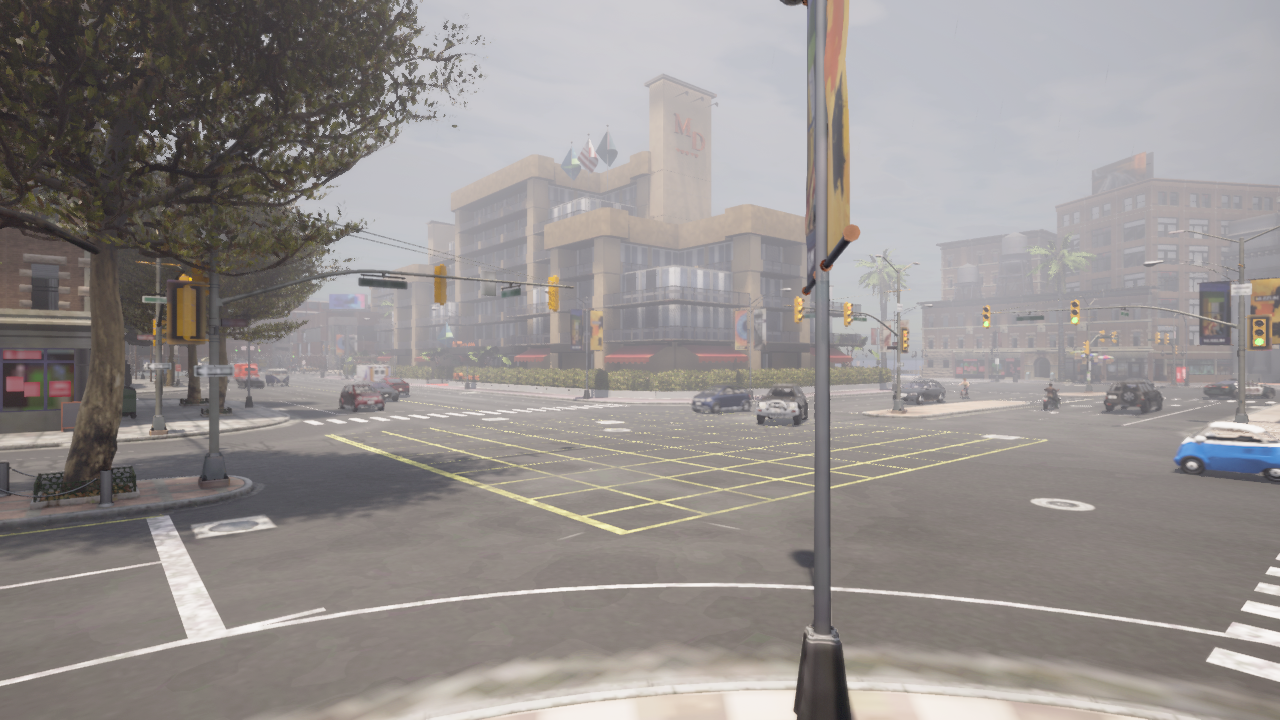}
    \end{minipage}
    \hfill
    \begin{minipage}{0.19\linewidth}
        \centering
        \includegraphics[width=\linewidth]{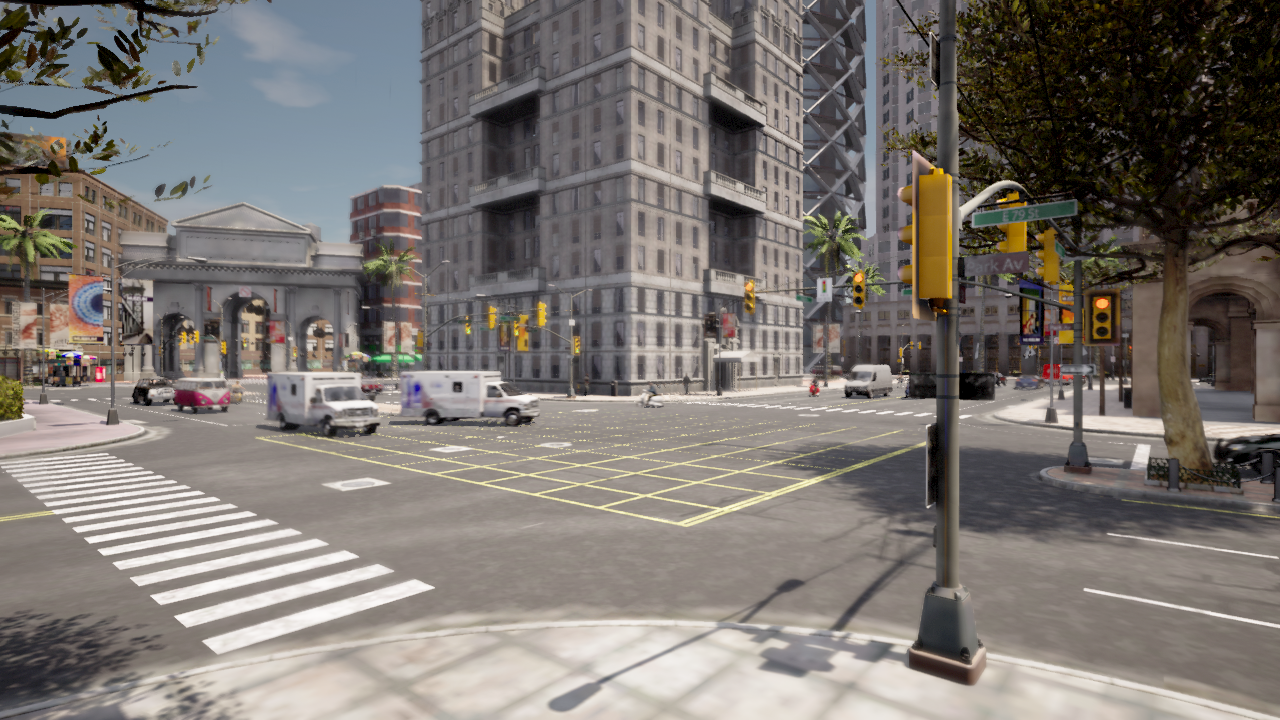}
    \end{minipage}
    \hfill
    \begin{minipage}{0.19\linewidth}
        \centering
        \includegraphics[width=\linewidth]{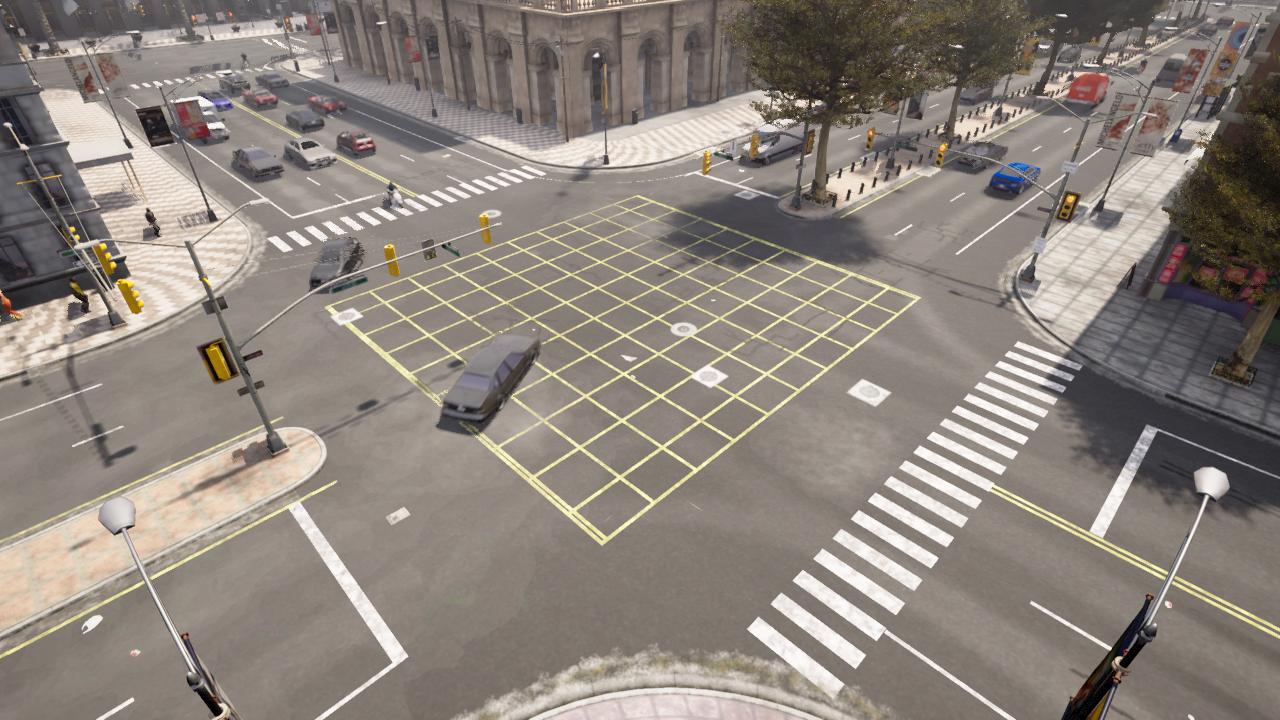}
    \end{minipage}
    \hfill
    \begin{minipage}{0.19\linewidth}
        \centering
        \includegraphics[width=\linewidth]{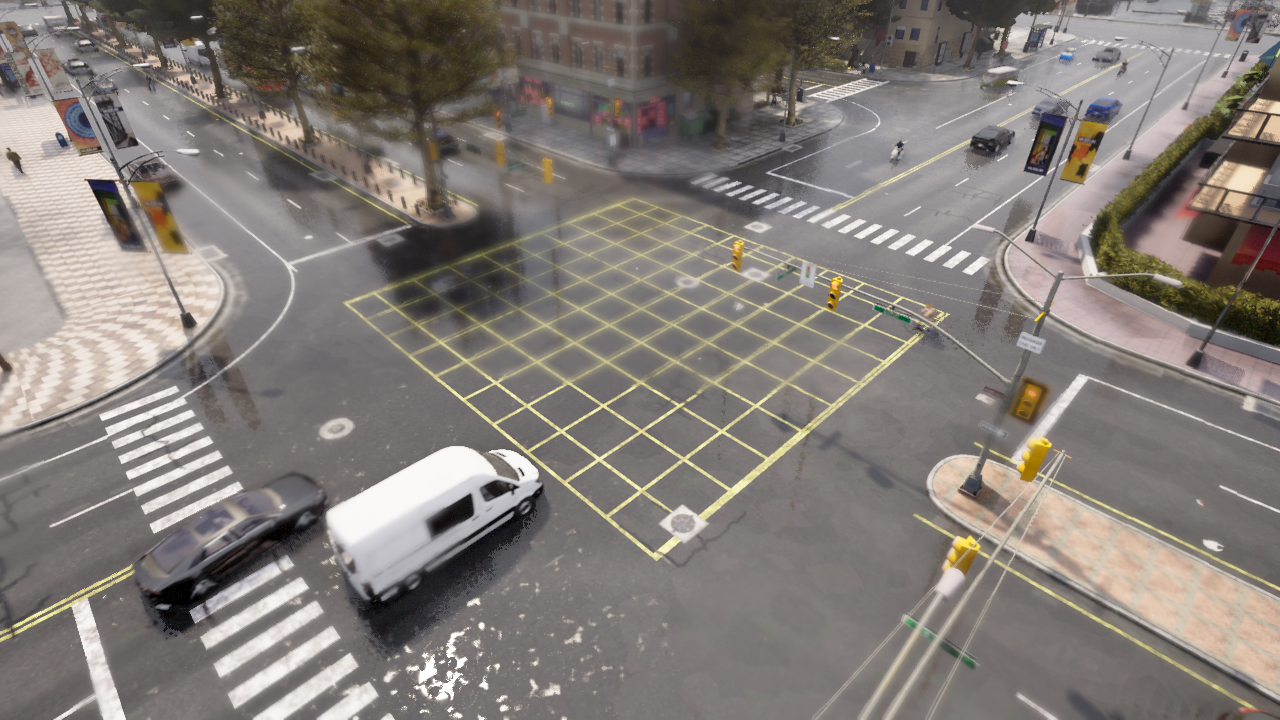}
    \end{minipage}
    \hfill
    \begin{minipage}{0.19\linewidth}
        \centering
        \includegraphics[width=\linewidth]{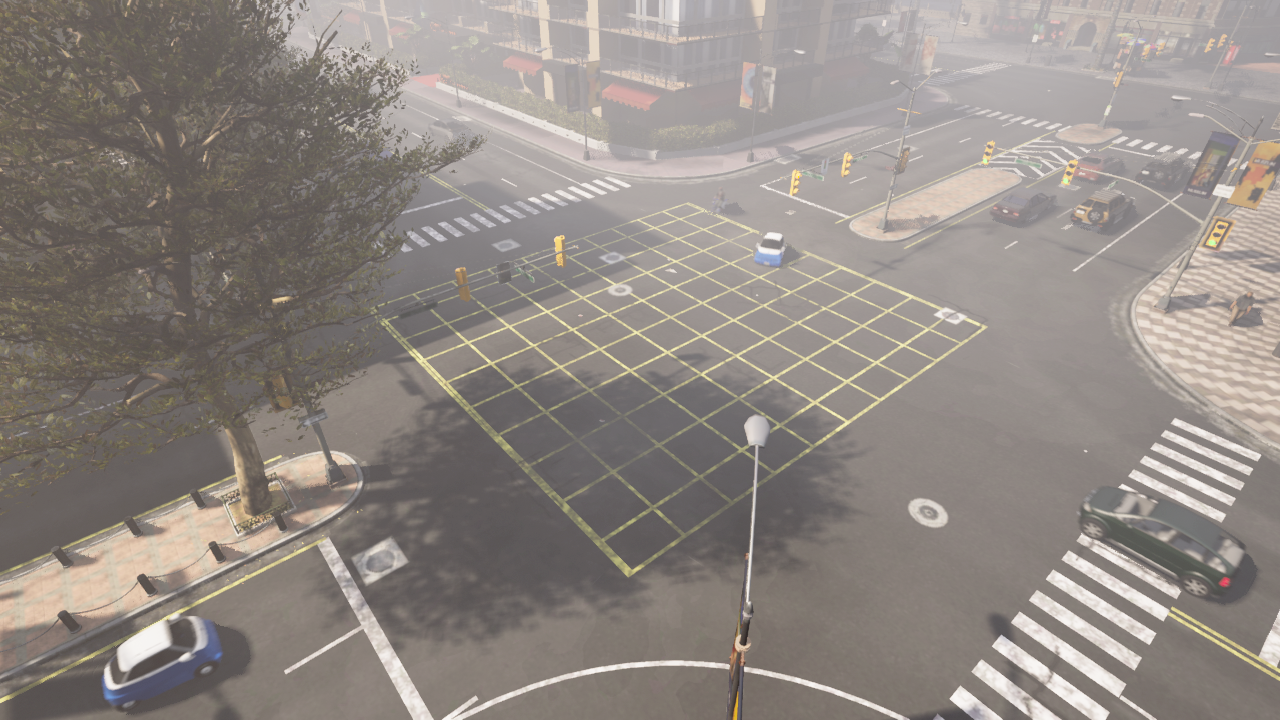}
    \end{minipage}

    \vspace{.5em}
    
    \begin{minipage}{0.19\linewidth}
        \centering
        \includegraphics[width=\linewidth]{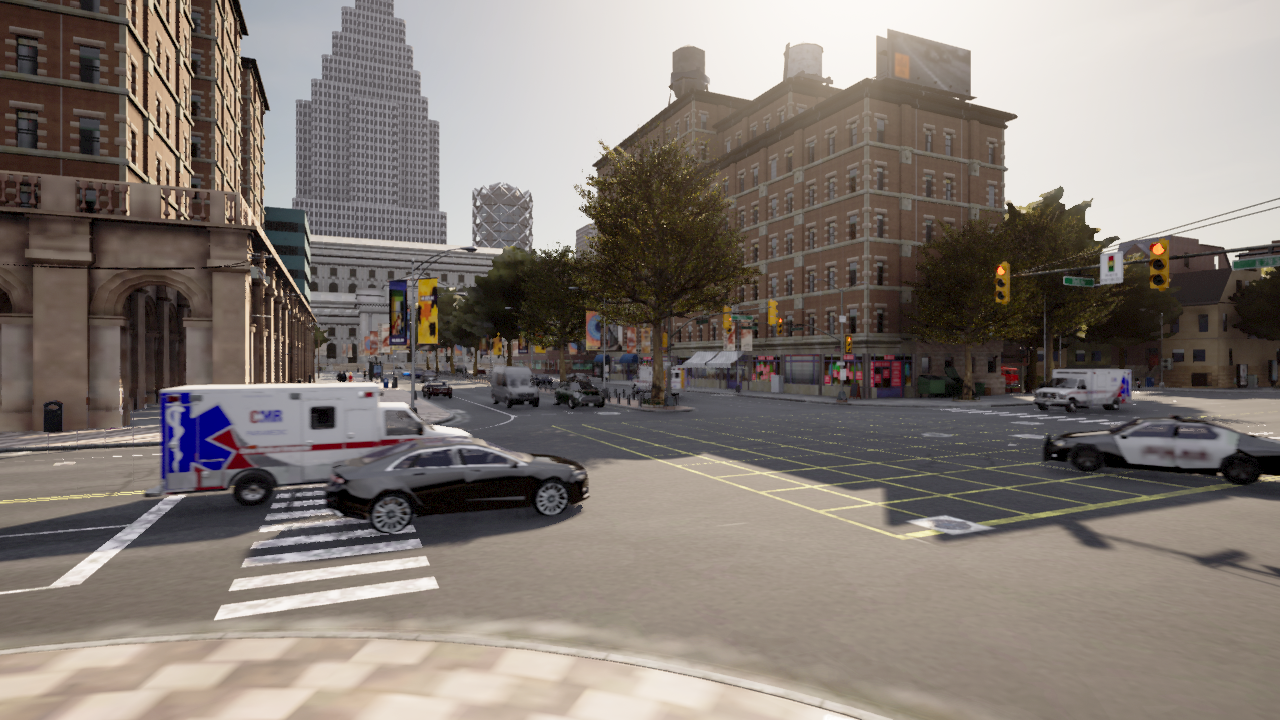}
    \end{minipage}
    \hfill
    \begin{minipage}{0.19\linewidth}
        \centering
        \includegraphics[width=\linewidth]{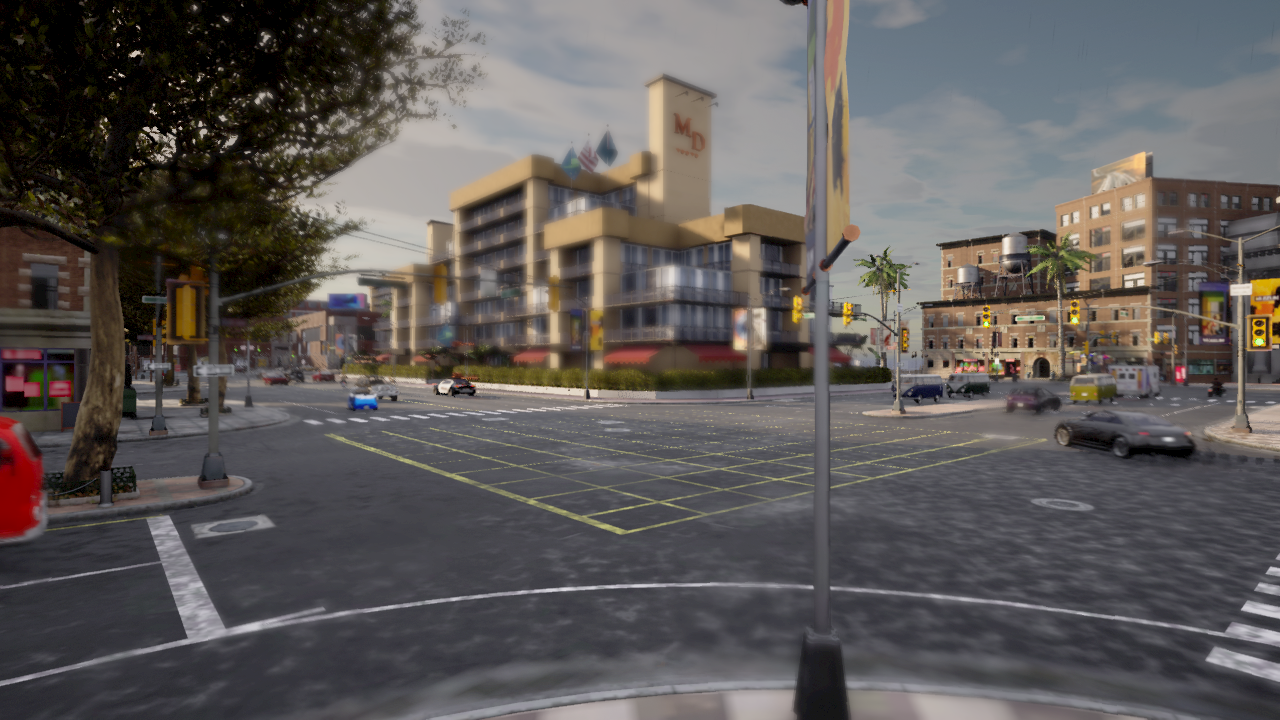}
    \end{minipage}
    \hfill
    \begin{minipage}{0.19\linewidth}
        \centering
        \includegraphics[width=\linewidth]{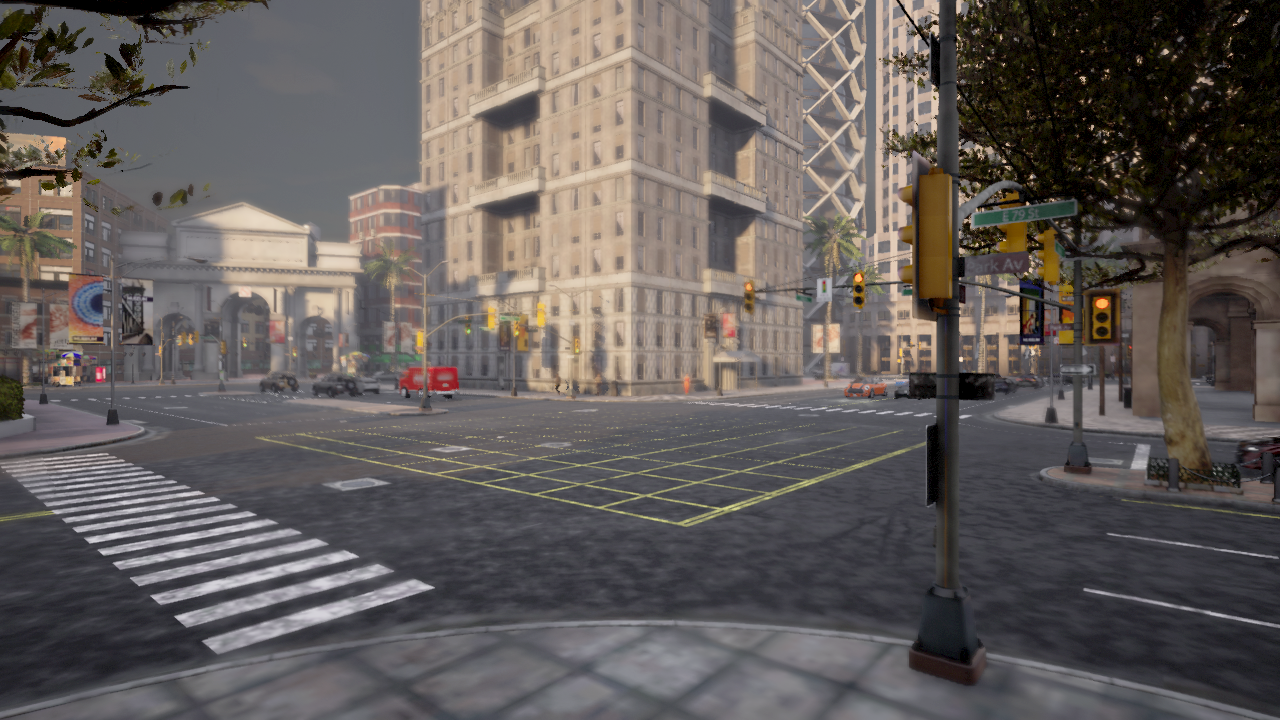}
    \end{minipage}
    \hfill
    \begin{minipage}{0.19\linewidth}
        \centering
        \includegraphics[width=\linewidth]{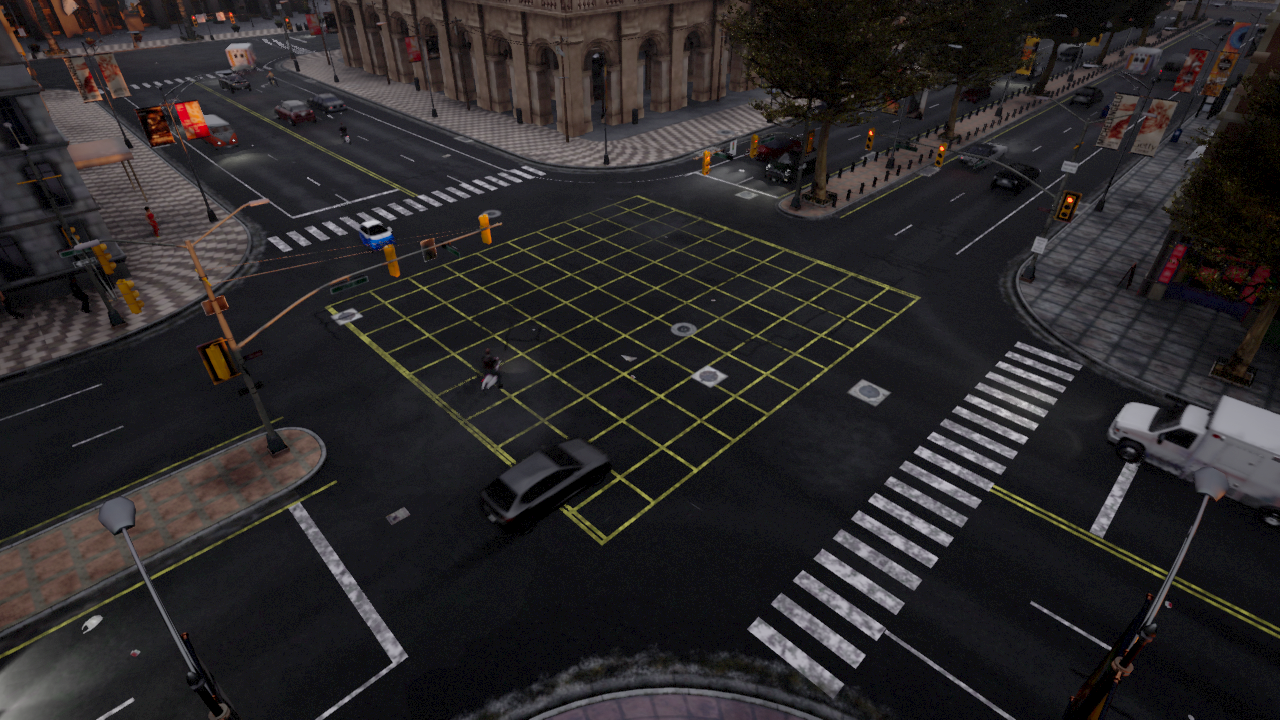}
    \end{minipage}
    \hfill
    \begin{minipage}{0.19\linewidth}
        \centering
        \includegraphics[width=\linewidth]{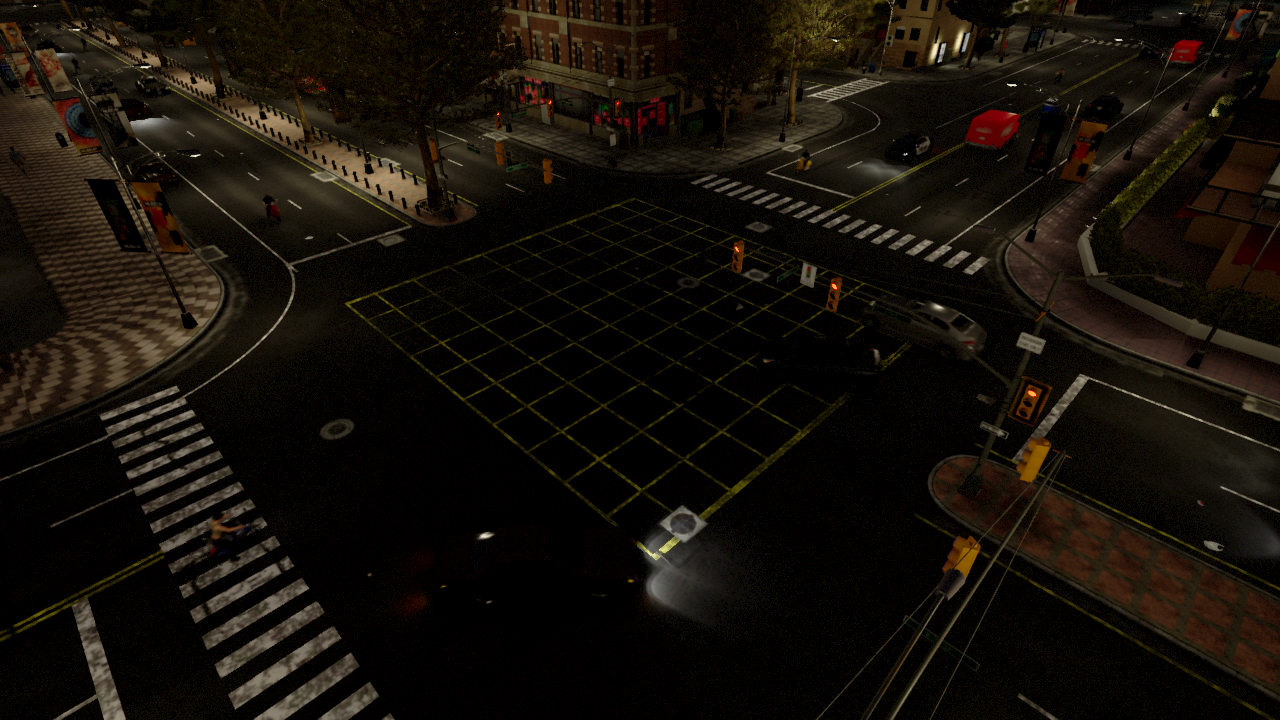}
    \end{minipage}
    
    \vspace{.5em}
    
    \begin{minipage}{0.19\linewidth}
        \centering
        \includegraphics[width=\linewidth]{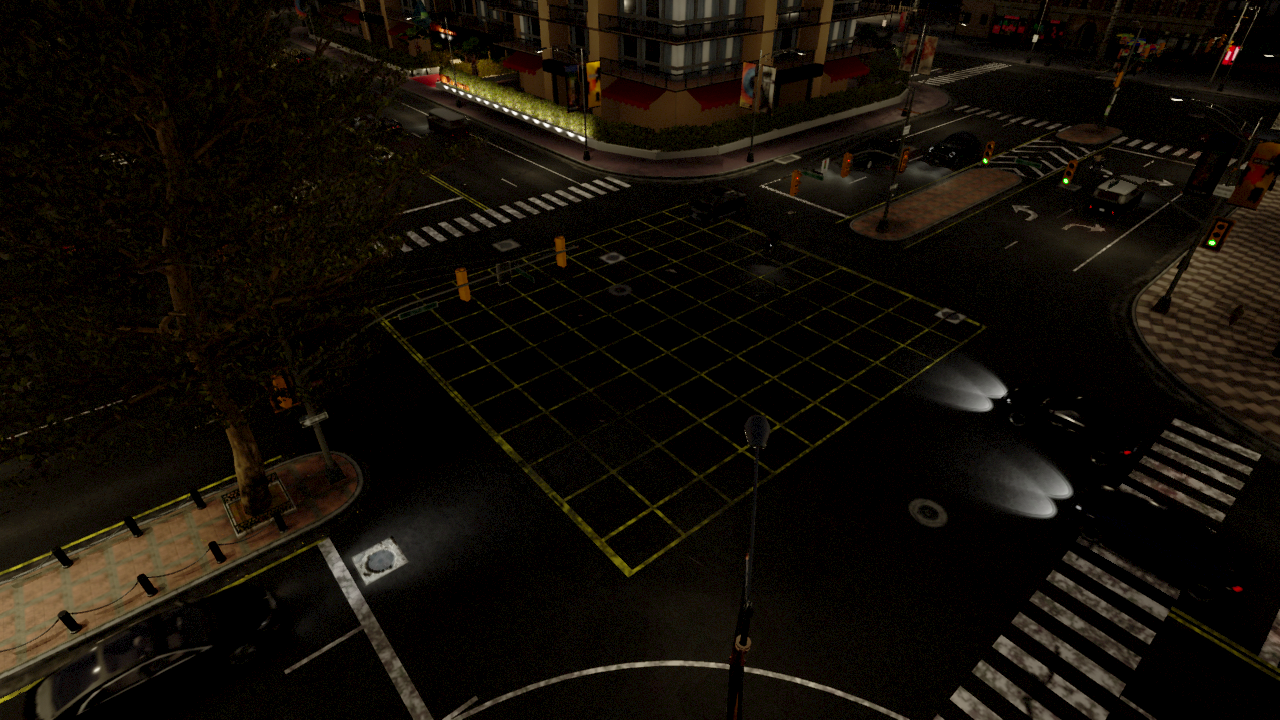}
    \end{minipage}
    \hfill
    \begin{minipage}{0.19\linewidth}
        \centering
        \includegraphics[width=\linewidth]{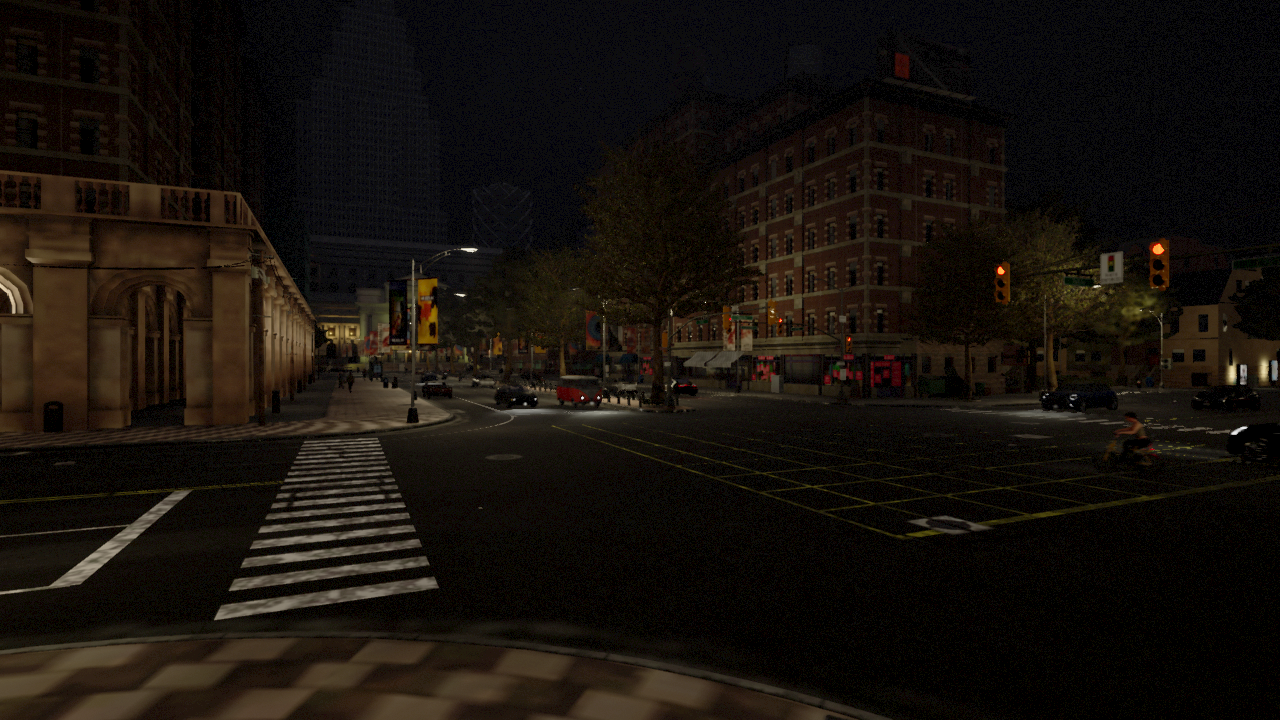}
    \end{minipage}
    \hfill
    \begin{minipage}{0.19\linewidth}
        \centering
        \includegraphics[width=\linewidth]{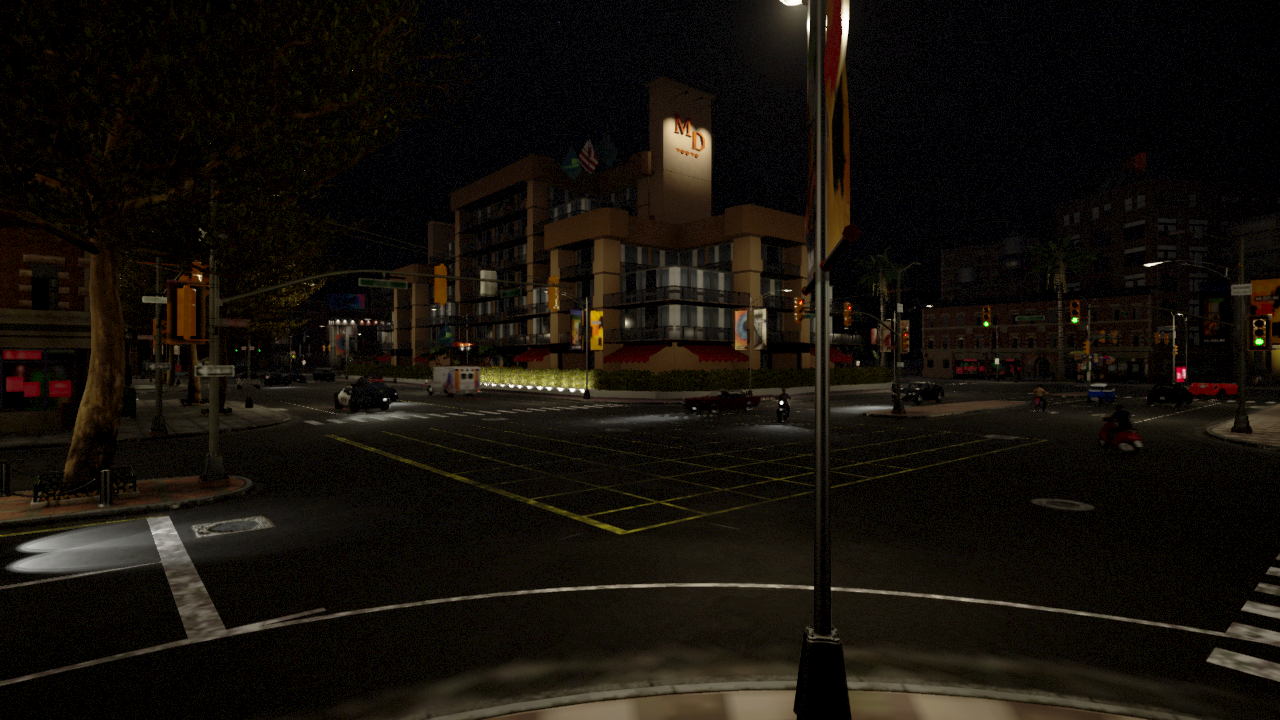}
    \end{minipage}
    \hfill
    \begin{minipage}{0.19\linewidth}
        \centering
        \includegraphics[width=\linewidth]{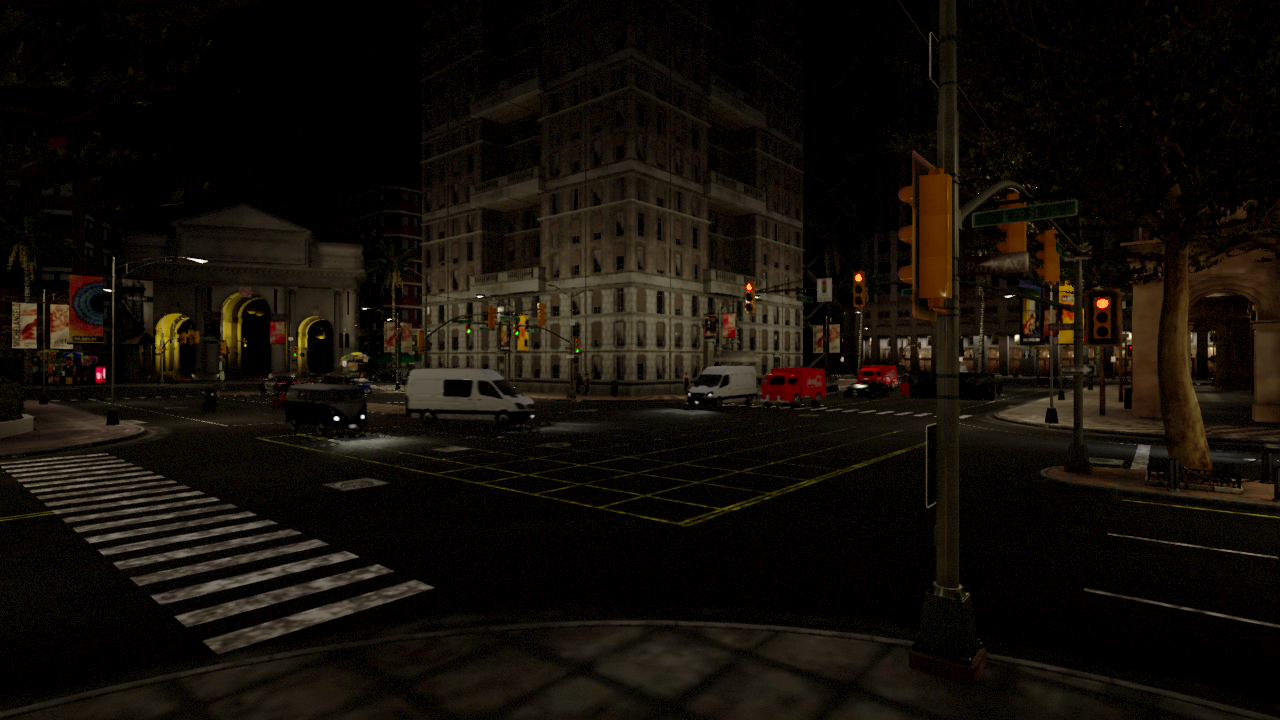}
    \end{minipage}
    \hfill
    \begin{minipage}{0.19\linewidth}
        \centering
        \includegraphics[width=\linewidth]{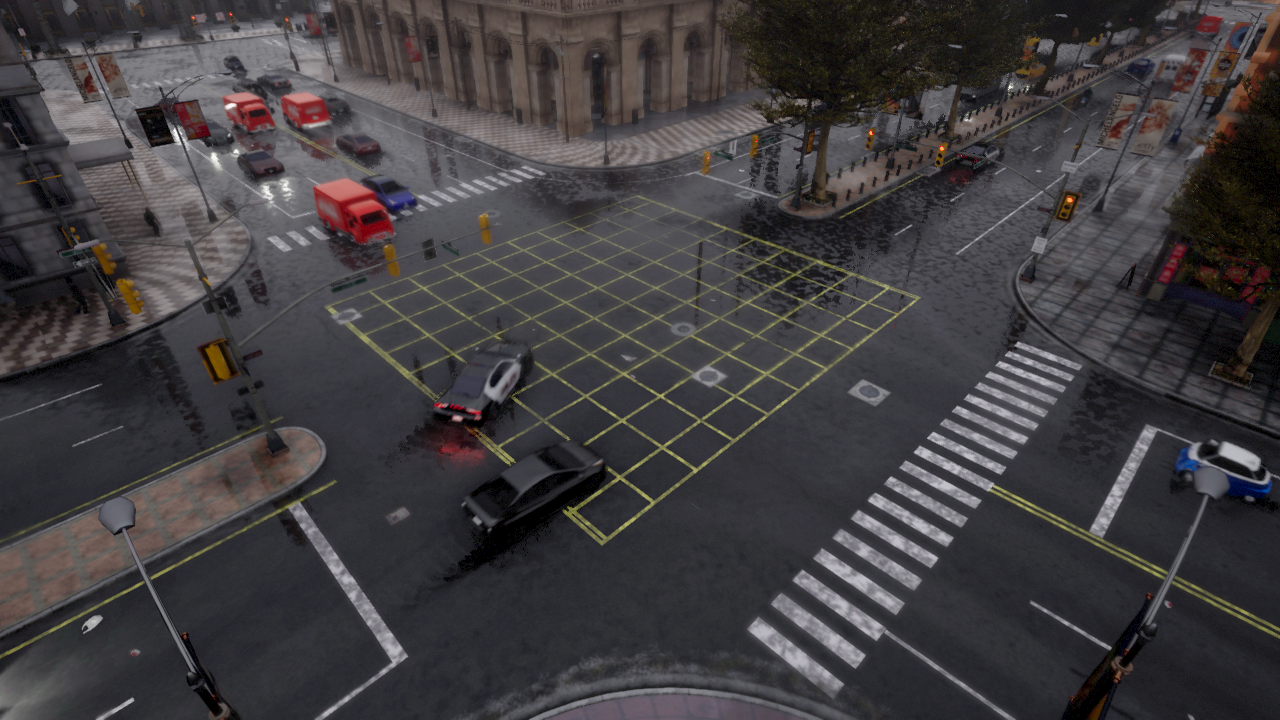}
    \end{minipage}
    
    \caption{First RGB frame of each of the $25$ video sequences of the \datasetAnaisAcronym dataset. This illustrates the various view points, illumination, and weather conditions across the different sequences.}
    \label{fig:rgb_sequences}
\end{figure*}

%% file: sec/4_performance.tex
\section{Performance}\label{sec:performance}

Evaluating a method capable of performing a task involves various aspects, ranging from its effectiveness in performing the task to the equipment used and the complexity of the method. This section addresses several of these aspects. 

\subsection{A Probabilistic Performance Evaluation Pipeline}
\label{sec:pipeline}

\input{sec/4_sub_performance.tex}

\subsection{Hardware Aspects}

Efficient deployment in edge and resource-constrained settings requires models with moderate memory demands and limited computational overhead. To evaluate these aspects, both the absolute memory usage and a normalized memory efficiency indicator are to be considered.

\paragraph{Peak Memory Usage} 
Memory requirements are quantified through the maximum CPU or GPU memory allocation observed during inference. This metric provides a direct measure of the resources required to execute a model and facilitates comparisons in terms of deployment feasibility on hardware platforms with different memory capacities.

\paragraph{Normalized Memory Score}
To express memory efficiency relative to a predefined budget, the peak memory consumption is further normalized. Given a target memory limit $\MEM_{target}$ and the measured peak usage $\MEM$, the corresponding score is computed as:
\begin{equation}
\MEM_{\Delta} = \max\left(0, \frac{\MEM}{\MEM_{target}} - 1\right)\point
\end{equation}
Lower values of $\MEM_\Delta$ indicate that the method operates closer to, or within, the desired memory constraints, whereas larger values reflect increasing deviations from the target budget.

\paragraph{Computational Complexity (\flops)}
The intrinsic computational cost of a model is measured by the number of floating-point operations (\flops) required for a single forward pass over a fixed-size input (\eg, a clip of $T$ frames at a given resolution). Unlike the processing frame rate, which depends on the target hardware and implementation, \flops provide a hardware-agnostic estimate of computational demand, allowing fair comparison across architectures regardless of the GPU or accelerator used for benchmarking. 

\paragraph{Number of Parameters and Model Size}
Model size is quantified through the total number of trainable parameters $N_{\text{params}}$, independently of the numerical precision or hardware used at inference. This metric offers an architecture-agnostic indication of storage footprint and complements the memory metrics above, which instead reflect runtime allocation; it can be expressed either as the total parameter count or as the model size in MB, where the latter depends on the data type used to store the parameters ---not limited to the floating-point formats FP16/FP32, but potentially a lower-precision type resulting from quantization.

\paragraph{Power Consumption}
Beyond memory footprint and computational complexity, the electrical power drawn during inference is a critical factor for battery-powered and always-on deployments (\eg, embedded GPUs or edge accelerators). This metric is quantified through the average power, expressed in Watts, sustained during steady-state inference on the target hardware platform. Unlike \flops, which is hardware-agnostic, measured power depends on the underlying accelerator and its energy efficiency for the executed operations, making it a complementary, hardware-aware counterpart to the computational complexity metric above.

\paragraph{Type of Architecture}
Numerical indicators alone do not fully capture a methods suitability for a given deployment scenario, since architectures with comparable memory footprints or computational complexity can differ substantially in their actual hardware behavior (\eg, parallelizability, exploitation of input sparsity, latency). One has therefore to report, for each method, a qualitative taxonomy of its dominant computational paradigm: 
\begin{itemize}
\item Convolutional Neural Networks (2D/3D CNNs).
\item Recurrent Neural Networks (LSTM/GRU, and related sequence models).
\item Attention-based/Transformers.
\item Graph Neural Networks (GNNs),\item Hybrid/multimodal architectures combining several of the above.
\item Classical, non-learning-based pipelines.
\end{itemize}
This categorical axis is intended to be reported alongside the quantitative metrics in the results tables.

\subsection{Real-time Performance and Deployability}

In surveillance applications, detection systems must not only be accurate but also capable of generating timely notifications while sustaining sufficient processing throughput. The following metrics are therefore adopted to evaluate real-time performance.

\paragraph{Notification delay}
This metric measures the latency between the onset of an event and its detection by the model. For each correctly identified positive event, the delay is defined as
$d_i = |p_i - g_i|$,
where $p_i$ and $g_i$ denote the predicted and ground-truth event onset times, respectively. The average delay $\delay$ over all true positive samples is then normalized according to
\begin{equation}
\delay_{norm} = \max\left(0, 1 - \frac{\delay}{T_{\max}}\right) \comma
\end{equation}
where $T_{\max}$ is the maximum acceptable delay for the task of interest, and higher values correspond to more prompt detections.

\paragraph{Processing frame rate}
Computational throughput is evaluated in terms of the average number of video frames processed per second on the target GPU. Let $\PFR$ denote the measured processing frame rate and $\PFR_{target}$ a reference throughput value. The associated normalized score is defined as:
\begin{equation}
\PFR_\Delta = \max\left(0, \frac{{\PFR}_{target}}{\PFR} - 1\right) \comma
\end{equation}
Higher scores of $\PFR_\Delta$ indicate more efficient processing and a greater ability to satisfy real-time constraints.

%% file: sec/4_sub_performance.tex
In this section, we present a complete and generic pipeline for evaluating a method performing an event detection task, analyzing its performance, and comparing it with others. This pipeline consists of four steps, as illustrated in \cref{fig:pipeline}. 
\input{figs/pipeline}
These steps are detailed hereafter. 

\subsubsection{Step 1: Evaluation}
\label{sec:pipeline:evaluation}
The first step of our pipeline is the evaluation, \ie the passage from an event detection method to its performance. We propose to cast detection problems as two-class crisp classification problems. 

\paragraph{Probabilistic approach}
It turns out that several well-known scores are often interpreted probabilistically. For example, we want to be able to interpret 
the accuracy as the probability to take the right decision, 
the precision as the probability that a positive decision is correct (or equivalently, as quoted by  \citet{Goutte2005AProbabilistic}, ``the probability that an object is relevant given that it is returned by the system''), 
the recall as the probability to take the right decision for a positive case (or equivalently as ``the probability that a relevant object is returned''), 
or the intersection-over-union as the probability for a decision to be correct given that either the ground-truth class or the predicted class is positive. 
Such probabilistic interpretations cannot be done rigorously without anchoring the concept of performance in probability theory. 
To this aim, we follow the framework of~\citet{Pierard2025Foundations} and 
present, hereafter, a general approach to specifying event detection evaluations in such a way as to obtain probabilistic performances for the outcomes $\outTN$ (true negative), $\outFP$ (false positive), $\outFN$ (false negative), and $\outTP$ (true positive). In this framework, \textit{a performance $P$ is modeled as a probability measure} on the measurable space $(\Omega,\Sigma)$, where $\Omega = \{\outTN,\outFP,\outFN,\outTP\}$ is the sample space or universe and $\Sigma = 2^\Omega$ is the event space (\ie, a sigma-algebra on $\Omega$).

\paragraph{Scores}
Scores are real functions of the performances. Unconditional probabilistic scores include 
the class priors:
\begin{align}
    \nprior(P) & = P(\{\outTN,\outFP\}) & \pprior(P) & = P(\{\outFN,\outTP\})
    \comma
\end{align}
the prediction rates:
\begin{align}
    \nrate(P) & = P(\{\outTN,\outFN\}) & \prate(P) & = P(\{\outFP,\outTP\})
    \comma
\end{align}
and the accuracy:
\begin{equation}
    \accuracy(P)
    = P(\{\outTN,\outTP\})
    \point
\end{equation}
Conditional probabilistic scores include 
the precision (\aka positive predictive value):
\begin{equation}
    \precision(P)
    = \ppv(P)
    = P(\{\outTP\}\vert\{\outFP,\outTP\})
    \comma
\end{equation}
the inverse precision (\aka negative predictive value):
\begin{equation}
    \npv(P)
    = P(\{\outTN\}\vert\{\outFN,\outTN\})
    \comma
\end{equation}
the recall (\aka true positive rate):
\begin{equation}
    \recall(P)
    = \tpr(P)
    = P(\{\outTP\}\vert\{\outFN,\outTP\})
    \comma
\end{equation}
the inverse recall (\aka true negative rate):
\begin{equation}
    \tnr(P)
    = P(\{\outTN\}\vert\{\outFP,\outTN\})
    \comma
\end{equation}
and the intersection over union (\aka Jaccard index~\cite{Jaccard1908Nouvelles}):
\begin{equation}
    \iou(P)
    = P(\{\outTP\}\vert\{\outFP,\outFN,\outTP\})
    \point
\end{equation}
It is also possible to use non-probabilistic scores such as the F-scores~\cite{vanRijsbergen1979Information,Christen2023AReview}:
\begin{equation}
    \fbeta(P)
    = \left(
        \frac{1}{1+\beta^2} \precision^{-1}(P)
        + \frac{\beta^2}{1+\beta^2} \recall^{-1}(P)
    \right)^{-1}
    \point
\end{equation}
For $\beta=1$, we have $\fone=\frac{2 \iou}{1+\iou}$. The balanced accuracy is
\begin{equation}
    \ba(P)
    = \frac12 \tnr(P) + \frac12 \tpr(P)
    \point
\end{equation}
As these last two examples show, even if a score cannot be expressed as the probability of some event, some scores may be expressed based on probabilistic scores. Thus, a rigorous probabilistic framework does not limit the scores to be probabilistic. 
Dozens of further measures or \emph{scores}, as we call them, have been proposed and compared~\cite{Canbek2022PToPI,Tharwat2021Classification,Powers2020Evaluation-arxiv}.
\citet{Berrar2019Performance} provides an overview of the fundamental performance measures for binary classification.
\citet{Sokolova2009ASystematic} systematically analyze 24 performance measures.
\citet{Canbek2022PToPI} compile a comprehensive table of 69 performance scoring strategies.
Further analyses and comparisons can be found in~\cite{Tharwat2021Classification,Powers2020Evaluation-arxiv,Parker2011AnAnalysis,Ferri2009AnExperimental}.
Finally, \citet{Christen2023AReview} provide a critical review specifically of the $\fone$ score and its variants. One should keep in mind that there is no such thing as one score being better than another in general. We will see in \cref{sec:pipeline:importance,sec:pipeline:ranking} that the infinite family of ranking scores~\cite{Pierard2025Foundations} are suitable when the final objective is to be able to compare event detection methods and to determine the best methods.

\paragraph{Numerical Representations of the Performance}
A probabilistic two-class crisp classification performance $P$ can be represented in different ways. A convenient representation is by a $2\times2$ confusion matrix (also called a contingency table) normalized so that the sum of all its elements equals to $1$. The elements of this matrix are thus the elementary probabilities $P(\{\outTN\})$, $P(\{\outFP\})$, $P(\{\outFN\})$, and $P(\{\outTP\})$; such matrices are drawn in \cref{fig:pipeline}. Alternatively, the performance can also be represented by a triplet of probability values such as $(\pprior,\prate,\accuracy)$ (which involves only unconditional probabilistic scores) or $(\pprior,\tnr,\tpr)$ (which involves both unconditional and conditional probabilistic scores). However, we will see in the second step (\cref{sec:pipeline:summarization}) that some numeric representations are more convenient than others when it comes to performing computations on them. In fact, the set of all possible performances is a $3$-dimensional simplex, \ie, a regular tetrahedron. This is why we need at least $3$ scores to obtain a continuous passage from the performance to its representation and a continuous inverse (the ``invariance of domain'' theorem \cite{Brouwer1911Beweis,Brouwer1912ZurInvarianz} %
implies that there is no continuous bijection whose inverse is continuous --no homeomorphism-- between $\mathcal{R}^n$ and $\mathcal{R}^m$ when $n \ne m$). That being said, not all triplets of scores can be used. In particular, it is useful to note that the often-reported triplet $(\precision, \recall, \fone)$ is insufficient to describe a performance. 

\paragraph{The random evaluation experiment}
The passage from any detection method to its performance can be rigorously specified by an elementary randomized test that leads to an outcome in $\Omega$ and that makes use of the detection method. We call it the \emph{random evaluation experiment}. It is this random (thought) experiment that gives the precise meaning to $\outTN$, $\outFP$, $\outFN$, and $\outTP$. Let us give three examples specific to the detection of events in videos.
\begin{experiment}[Evaluation of background subtraction methods]
    Consider a video clip (data source). Apply an oracle (which could be a human expert) to decide for every pixel of every frame whether it belongs to the background or to the foreground. Apply the background subtraction method (that has to be evaluated) on the clip. Choose a frame and a pixel at random and look at the corresponding outputs $y$ and $\hat{y}$ of the oracle and method, respectively. If $y=\hat{y}=background$, return $\outTN$. If $y=background$ and $\hat{y}=foreground$, return $\outFP$. If $y=foreground$ and $\hat{y}=background$, return $\outFN$. Otherwise, %
    return $\outTP$.
\end{experiment}
\begin{experiment}[Evaluation of pedestrian detectors]
    Consider a video clip (data source). Choose a frame at random. Apply an oracle (which could be a human expert) to obtain the set $y$ of ground-truth bounding boxes around all pedestrians. Apply the detector to obtain the set $\hat{y}$ of predicted bounding boxes. If $y=\hat{y}=\emptyset$, return $\outTN$. Otherwise, apply an arbitrary but well-specified matching criteria that associates at most an element of $y$ to every element of $\hat{y}$ and at most an element of $\hat{y}$ to every element of $y$. Choose an element at random in $y\cup\hat{y}$. If it belongs to $\hat{y}$ and has not been associated, return $\outFP$. If it belongs to $y$ and has not been associated, return $\outFN$. Finally, if it is associated, return $\outTP$.
\end{experiment}
\begin{experiment}[Evaluation of geographic feature detectors]
    Consider an image obtained by remote sensing (\eg, an aerial or satellite photo) on which some geographic features (\eg, trees or buildings) have to be detected, , like the elevation magnitude. Apply an oracle (which could be a human expert) to obtain a map of real values specifying the ground-truth magnitude in each pixel. Apply the detector to obtain the estimated map. Consider also some arbitrarily chosen real-valued probability distribution $T$, and draw a threshold $t$ at random following $T$. Choose a pixel at random and retrieve the ground-truth and estimated magnitude values, respectively $y$ and $\hat{y}$, in this pixel. Return 
    $\outTN$ if $\max(y,\hat{y})<t$ and $\outTP$ if $t<\min(y,\hat{y})$. In the other cases, return $\outFP$ if $y<\hat{y}$ and $\outFN$ otherwise.
\end{experiment}
\noindent This last example is inspired by the work of \citet{Krasnodebska2025Advancing} who defined ``continuous versions'' of $\precision$, $\recall$, $\fone$, and $\iou$. For magnitudes in $[0,M]$, if we take for $T$ a uniform distribution over the range $[0,M]$, then the values of the continuous scores of \cite{Krasnodebska2025Advancing} are the same as the values taken by the classical scores $\precision$, $\recall$, $\fone$, and $\iou$ for the performance resulting from this random evaluation experiment. The approach presented here is a generalization, as there is no notion of true negative in the work of \citet{Krasnodebska2025Advancing}. With the approach presented here, any score defined for two-class crisp classification could be used.

As these examples show, we have a great deal of freedom in choosing the experience. However, we will always be fully satisfied when the evaluation outcome is $\outTN$ or $\outTP$, and not satisfied when it is $\outFP$ or $\outFN$. Note that some results presented hereafter are particularizations of more general results from~\cite{Pierard2025Foundations} in the case of a binary satisfaction.

\paragraph{Practical implementation}
One should make the distinction between such a random (thought) evaluation experiment and the implemented evaluation algorithm. Indeed, the performance could be estimated by implementing the chosen random evaluation experiment and running it a high number of times (Monte-Carlo approach). However, in practice, it is much more common and convenient to assume that the evaluated method is deterministic, to run it once, and to compute the probabilities by counting the number of cases for which $\outTN$, $\outFP$, $\outFN$, and $\outTP$ are chosen as outcomes. What matters is the consistency between the specified random experiment and the implemented evaluation algorithm. Moreover, in practice, we can assume the frequentist interpretation of probabilities~\cite{Neyman1977Frequentist}, which requires a sufficient number of samples to be statistically meaningful.
Any probability can then be estimated as the relative frequency of a given outcome over all cases.
For example, accuracy is estimated by the ratio $(\TP+\TN)/(\TP+\TN+\FP+\FN)$, where $\TP$ denotes the observed count of true positives, $\TN$ of true negatives, \etc.

\subsubsection{Step 2: Summarization}
\label{sec:pipeline:summarization}
Evaluating event detection methods on a single data source (individual video clip, scene, or evaluation domain) only gives a narrow view of the method's performance. 
In practice, it is therefore common to apply a method to several videos and evaluate the average performance. 

\paragraph{Averaging performances instead of score values}
As a starting point, it should be emphasized that averaging performances differs from averaging score values. Averaging score values is inadequate, as the resulting mean values are uninterpretable. 
To see this, it suffices to note that, for a given set of scores, the mean values do not correspond to any possible performance. To give a concrete example, despite \citet{Davis2006TheRelationship} showing that there is a bijection between the $ROC=(1-\tnr,\tpr,\pprior)$ (Receiver Operating Characteristic) and $PR=(\recall,\precision,\pprior)$ (Precision-Recall) spaces, \citet{Pierard2020Summarizing} pointed out that the centroid in $ROC$ does not correspond to the centroid in $PR$. As a second example, the centroid in $PR$ can fall into the unachievable region reported by \citet{Boyd2012Unachievable} for the $PR$ space. In general, mean values are inconsistent across different scores.
As a solution, \citet{Pierard2020Summarizing} proposed the following probabilistically grounded \emph{summarization} principle to average the performances $P$. Let us assume that some given detection method has been evaluated for several data sources and that we need to average the resulting performances. If the evaluation experiments (see \cref{sec:pipeline:evaluation}) differ only by the data source, then we can also consider (by thought) evaluating with a hybrid source such that drawing some data at random from it (\eg, a frame or a pixel) is equivalent to drawing a base source (\eg, a video) and then drawing the data at random from it. More explicitly, we can consider a mixture of the distributions of data on which the detector was previously evaluated, \eg, a mixture of videos. Indeed, the mixture weights (\ie, the weights given to the various videos) can be chosen arbitrarily. The idea is then to define the \emph{summarized (averaged) performance} as the performance resulting from the evaluation with this hybrid data source~\cite{Pierard2025Multidomain-arxiv}.

\paragraph{What the evaluation is telling us}
It turns out that some random evaluation experiments are \emph{linear with respect to the data source} in the sense that the performance resulting from a mixture of sources is the convex combination of the performances with the weights of the combination corresponding to the chosen mixture weights. %
For such evaluations, the summarized performance for a method $m$ is thus
\begin{equation}
    P_{summarized} = \sum_s \lambda_s P_s
    \quad
    \textrm{with}
    \quad
    \begin{cases}
        \lambda_s \ge 0\\
        \sum_s \lambda_s = 1
    \end{cases}
    \comma
\end{equation}
where $P_s$ is the performance of the method $m$ for the $s$\textsuperscript{th} data source and $\lambda_s$ denotes the relative weight given to this source. 
The summarized confusion matrix is just the weighted arithmetic mean of those for the performances that need to be averaged. More generally, for such evaluations, one can arithmetically average any numerical representations of performances involving only linear scores (unconditional probabilistic scores or expected value scores). This shows a direct benefit, apart from avoiding Bertrand's paradox~\cite{Bertrand1889Calcul}, of explicitly specifying the random evaluation experiment: we can then determine its properties and derive meaningful operations (\eg, an averaging) on performances based on them.
Summarization provides a theoretical justification for taking the arithmetic mean of the confusion matrices, but one cannot simply take the arithmetic mean of any numerical representation. For example, as the class priors are video-dependent, scores like the true negative and positive rates are not linear with respect to the elements of the confusion matrix. Therefore, it is not correct to arithmetically average numerical representations based on them, such as $(\pprior,\tnr,\tpr)$.

\paragraph{The implication for scores}
While it is easier to start by averaging performances before applying the score that we want, it is sometimes also possible to work directly on scores, but the formulas to do so are not intuitive. The formulas for unconditional and conditional probabilistic scores can be found in~\cite{Pierard2020Summarizing}. 
The formulas for ranking scores can be found in~\cite{Pierard2025MultiDomain}. To give a concrete example, consider a collection $\mathbb{V}$ of videos, and let us denote the weight given to the video $v \in \mathbb{V}$ by $\lambda_v$, the performance on the video $v$ by $P_v$, and the summarized performance by $P_\mathbb{V}$. The resulting summary positive prior is $\pprior(P_\mathbb{V}) = \sum_{v \in \mathbb{V}} \lambda_v\,\pprior(P_v)$, and the summarized \tpr is
\begin{equation}\label{eq:sumTPR}
  \tpr(P_\mathbb{V})
    = \frac{\sum_{v\in\mathbb{V}}
            \lambda_v\,\pprior(P_v)\,\tpr(P_v)}
           {\pprior(P_\mathbb{V})}\point
\end{equation}
This is a \emph{prior-weighted} mean of \tpr, not an arithmetic mean, because the positive-class count of each video provides the correct weight.

\subsubsection{Step 3: Importance-based Analysis}
\label{sec:pipeline:importance}
The third step of our pipeline aims at considering some application-specific preferences for the performance analysis. It is worth keeping in mind that, generally speaking, no single score is better than the others. It all depends on the application needs. So, in this step, we explore a wide range of scores for which one has a crystal-clear understanding of the corresponding preferences; this leads us to the notion of \emph{canonical ranking scores}~\cite{Pierard2024TheTile-arxiv,Halin2024AHitchhikers-arxiv}

\paragraph{Canonical ranking scores}
Let $a\in[0,1]$ be the relative importance given to the true positives \wrt the true negatives and $b\in[0,1]$ be the relative importance given to the false negatives \wrt the false positives. These can be chosen arbitrarily according to the preferences corresponding to the target applications. For tasks similar to two-class crisp classification, the canonical ranking scores are defined in~\cite{Pierard2024TheTile-arxiv,Halin2024AHitchhikers-arxiv} as
\begin{equation}
    \frac{
        (1-a) P(\{\outTN\}) + a P(\{\outTP\})
    }{
        (1-a) P(\{\outTN\}) + (1-b) P(\{\outFP\}) + b P(\{\outFN\}) + a P(\{\outTP\})
    }
    \point
\end{equation}
Our motivation for using this family of scores, that forms a continuum parameterized by $(a,b)$, is twofold. First, as will be discussed in details in \cref{sec:pipeline:ranking}, these scores are all suitable to induce meaningful rankings (hence their name). Second, this family of scores includes some well-known scores that have already been used in the process of evaluating methods dealing with the task of event detection. Thus, we obtain the accuracy $\accuracy$ with $(a,b)=(\frac12,\frac12)$. Likewise, we have $\precision=\ppv$ with $(a,b)=(1,0)$, $\npv$ with $(a,b)=(0,1)$, $\tpr$ with $(a,b)=(1,1)$, $\tnr$ with $(a,b)=(0,0)$, $\fbeta$ with $(a,b)=(1,\frac{\beta^2}{1+\beta^2})$, and $\ba$ with $(a,b)=(\nprior,\nprior)$.

\paragraph{Tiles}
Choosing the importance point $(a,b)$ to use is itself non-trivial and application-dependent.
The \emph{Tile}~\cite{Pierard2024TheTile-arxiv} is a graphical tool that solves this problem by displaying the entire 2-parameter family of ranking scores simultaneously on a unit square (see \cref{fig:tile}).
\input{figs/tile}
Note that, according to \citet{Mitchell2019Model}, the error types that can be derived from a confusion matrix are the false positive rate ($=1-\tnr$), false negative rate ($=1-\tpr$), false discovery rate ($=1-\ppv$), and false omission rate ($=1-\npv$). It is interesting to note that these scores correspond (if we take their complements) to the four corners of the Tile. Also, according to these authors, the relative importance of each of these error types is system-, product-, and context-dependent. This interpretation matches pretty well the concept of application-specific preferences that is used for the parameters $a$ and $b$ in the literature about Tiles~\cite{Pierard2024TheTile-arxiv,Halin2024AHitchhikers-arxiv,Pierard2025Multidomain-arxiv,Pierard2025AMethodology}.

\paragraph{Flavors}
The Tile can be used to display various types of information. For example, we can set a false-color image as the background of the Tile to depict in point $(a,b)$ the value taken by the corresponding canonical ranking score. This is the \emph{Value Tile}. The \emph{Sorbetto} library~\cite{Pierard2025Sorbetto-zenodo} can be used to draw Tiles with various flavors and annotations on top of them. In fact, the Tile domain serves as a canvas onto which diverse evaluation information can be projected; each such projection is called a \emph{flavor}~\cite{Halin2024AHitchhikers-arxiv}.
Some flavors introduced so far include~\cite{Halin2024AHitchhikers-arxiv}: (i)~the \emph{Value Tile}, which maps the score value achieved by a given method at every Tile point ---this answers ``how well does method~A score according to each possible ranking criterion?''; (ii)~the \emph{Entity Tile}, which maps the \emph{identity} of the best method for each ranking criterion ---answering ``which method wins under each possible criterion?''; (iii)~the \emph{Baseline Value Tile} and (iv)~the \emph{State-of-the-Art Tile}, showing the infimum and supremum of score values over a benchmark; and (v)~the \emph{Ranking Tile}, which displays the rank of a given method for every criterion.

\paragraph{Taking the chance into account}
Since the Finley ``affair'' \cite{Finley1884Tornado}, it is well known that ``raw'' scores can be misleading as one can obtain relatively high values by chance. Many papers proposed to correct the accuracy in different ways~\cite{Bennett1954Communications,Scott1955Reliability,Cohen1960ACoefficient,Appleman1960AFallacy}, and others introduced scores to measure the skill of classifiers \cite{Gilbert1884Finley,Peirce1884TheNumerical,Yule1900OnTheAssociation,Heidke1926Berechnung,Clayton1934Rating}. Unfortunately, the relationship between the resulting scores and the applicative preferences is unclear for most of them. The Tile offers another path. As depicted in pink on the Value Tiles of \cref{fig:pipeline}, we can hatch the areas in which the value of the canonical ranking score is less than either the value for the no-skill classifier predicting always the negative class or the value for the no-skill classifier predicting always the negative class. The nearest the performance is from the no-skill performances (those achievable only by chance, for which the ground truth and predicted classes are independent), the larger the hatched areas are. If the performance is no-skill or below them, then the entire Value Tile is hatched.

\paragraph{Comparison with older graphical tools}
In the context of this paper, we focus on comparing crisp classifiers rather than analyzing soft classifiers (which can be seen as families of crisp classifiers). Graphical analysis of performances has historically relied on the \emph{Receiver Operating Characteristic} (ROC) curve, which plots \tpr against $\fpr=1-\tnr$ (the false positive rate) for a continuously thresholded soft classifier. 
The ROC framework was born from signal detection theory, developed and first presented by \citet{Peterson1954TheTheory}, and its application to diagnostic settings was formalized later by \citet{Swets1988Measuring}.
The Precision--Recall (PR) space has been advocated as a complement when classes are imbalanced~\cite{Boyd2013Area,Davis2006TheRelationship}, but this comes with its drawbacks, such as the existence of an unreachable area~\cite{Boyd2012Unachievable}. 
Both spaces, however, describe a parametric curve of soft classifiers rather than the performance of a single operating point.
This is an advantage when characterizing a parametric family of classifiers, but impractical when comparing several distinct crisp classifiers. 
Moreover, these spaces were not designed to read score values easily (or to observe rankings, as will be needed in our last step, see \cref{sec:pipeline:ranking}). As a solution, \citet{Flach2003TheGeometry} proposed to represent scores in ROC for some arbitrarily chosen class priors through isometrics (the same could be done in the PR space). But even if this is a solution for a single score, this path is impractical when it comes to exploring a large family of scores. The Tile solves this issue: while performances are projected as points in the ROC and PR spaces, it is the applicative preferences and the corresponding canonical ranking scores that are projected as points in the Tile, so that one can have a global overview at a glance.

\subsubsection{Step 4: Ranking}
\label{sec:pipeline:ranking}
Ranking methods based on their performances is of crucial importance, not only for the organizers of challenges and competitions but also for those who need to select a method for their specific application. We assume that the performances are all comparable in the sense that they result from the same random evaluation experiment. While there is no universal ranking, not all ways of computing a ranking are suitable. Thus, our pipeline would not be complete without the description of how to derive meaningful rankings of event detection methods. For comparing crisp classifiers (\aka methods), the choice of adequate scores to rank classifiers is less straightforward than it may appear.
\citet{Pierard2025Foundations} introduced the first axiomatic framework for \emph{performance-based rankings} that is consistent with the probabilistic performance model that we chose in the first step (see \cref{sec:pipeline:evaluation}).

\paragraph{Stable and meaningful rankings}
Maybe the most obvious requirement is the stability regarding the insertion of any new method. The conclusions about which methods are better than others should not change over time, \eg, when a new method is inserted into the ranking. Looking back at what has been done in the change detection community over the last decade, it appears that this requirement was not always met. For example, the stability is \emph{not} guaranteed by the rank-aggregation procedure used in \CDNET~2014~\cite{Goyette2014ANovel}. Within any given category of videos, the methods were ranked with respect to a few arbitrarily chosen scores, and the overall ranking followed the average of the rank values across scores. Inserting a new competitor shifts all rank positions, so the mean-of-ranks of existing methods can change and their relative order can reverse even when their scores are unchanged. A similar problem occurs when it comes to computing an overall ranking for the complete dataset based on the per-category rankings. Unfortunately, this means that we cannot trust any conclusion about what methods are better than others that has been published in the scientific literature and that was obtained with that methodology. To obtain stable rankings, it is advised to start by choosing a unique ordering between all possible performances and then to derive the ranking of methods from this fixed ordering~\cite{Pierard2025Foundations}. A preorder between all possible performances is naturally induced by any score, but not all scores induce meaningful performance orderings. Before deriving a ranking from any arbitrarily chosen score, one has to prove that a method is worse than/equivalent to/better than another if and only if the value given by the score for the performance of the first is lower than/equal to/higher than the value for the second. This is less straightforward than it may appear.

\paragraph{What the evaluation is telling us} 
For any predefined list of methods, we can always consider the hybrid methods that start by selecting \emph{blindly} and at random (with arbitrary probabilities) one base method in this list before executing it. As underlined in~\cite{Pierard2025Foundations}, it would not make sense for such a hybrid method to be considered worse than the worst of the base methods or better than the best of the base methods. To ensure this, we need to look back once again at what is done in the evaluation step (see \cref{sec:pipeline:evaluation}). It turns out that some random evaluation experiments are \emph{linear with respect to the evaluated method} in the sense that the performance resulting from such a hybrid method is the convex combination of the performances resulting from the base methods, with the weights of the combination corresponding to the selection probabilities. In mathematical terms, 
\begin{equation}
    P_{hybrid} = \sum_m \lambda_m P_m
    \quad
    \textrm{with}
    \quad
    \begin{cases}
        \lambda_m \ge 0\\
        \sum_m \lambda_m = 1
    \end{cases}
    \comma
\end{equation}
where $P_m$ is the performance of the $m$\textsuperscript{th} method and $\lambda_m$ denotes the relative weight given to this method. 
This is the case, in particular, when the evaluated method is used only once in the random evaluation experiment. The linearity with respect to the evaluated method is sometimes implicitly assumed in the literature. We argue that it is risky not to mention that implicit assumption, as the conclusions that can be drawn from this are not true in general. To give a concrete example, in his famous paper providing an introduction to ROC analysis, \citet{Fawcett2006AnIntroduction} devoted a complete section to the interpolation of classifiers and explained that
\begin{align}
    \fpr(P_{hybrid}) = \sum_m \lambda_m \fpr(P_m) \comma \\
    \tpr(P_{hybrid}) = \sum_m \lambda_m \tpr(P_m) \point
\end{align}
This is an important result as it justifies connecting points with line segments in ROC (and thus with curves in PR, since the passage between ROC and PR is non-linear~\cite{Davis2006TheRelationship}) to obtain continuous ``ROC curves''. The interpolation of \citet{Fawcett2006AnIntroduction} is nevertheless valid under some common conditions, \eg when the random evaluation experiment is linear with respect to the evaluated method and when the class priors are the same for all combined performances.

\paragraph{Ranking scores}
Let us now come back to our main concern, which is to obtain meaningful rankings. It has been proven in~\cite{Pierard2025Foundations} that all performance orderings induced by \emph{ranking scores} can safely be used to rank when the evaluation is linear with respect to the evaluated method. This family is parameterized by an \emph{importance function} $I:\Omega\to\mathbb{R}_{+}$ that assigns a non-negative weight to each of the four outcomes~\cite{Halin2024AHitchhikers-arxiv}:
\begin{equation}\label{eq:ranking}
  R_I(P)
    = \frac{I(\outTN)\,P(\{\outTN\}) + I(\outTP)\,P(\{\outTP\})}
           {\sum_{\omega\in\Omega} I(\omega)\,P(\{\omega\})} \point
\end{equation}
Every score in this family rewards correct outcomes (\outTN, \outTP) relative to all outcomes, weighted by how much each type of error or success matters to the application. For all importances $I$, $R_I(P)=1$ for the best performances (\ie, when $P(\{tn,tp\})=1$) and $R_I(P)=0$ for the worst performances (\ie, when $P(\{tn,tp\})=0$). Moreover, under the assumption of a linear random evaluation with respect to the evaluated method, for all hybrid methods, $\min_m R_I(P_m) \le R_I(P_{hybrid})\le \max_m R_I(P_m)$. The conclusion drawn from that by \citet{Pierard2025Foundations} is that we can say that a method is worse than/equivalent to/better than another if and only if the value given by the ranking score for the performance of the first is lower than/equal to/higher than the value for the second. However, we emphasize that the various $I$ can lead to different performance orderings, and thus to different rankings. 
The canonical ranking scores we just mentioned in \cref{sec:pipeline:importance} form a subset of the larger family of ranking scores. A ranking score is a canonical ranking score if and only if $I(\outTN)+I(\outTP)=I(\outFP)+I(\outFN)$. It has been established in~\cite{Pierard2025Foundations} that several ranking scores can lead to the same performance ordering and thus to the same rankings. More precisely, two ranking scores $R_{I_1}$ and $R_{I_2}$ induce the same performance ordering when $\TileA(I_1)=\TileA(I_2)$ and $\TileB(I_1)=\TileB(I_2)$ where $\TileA(I) = I(\outTP)/(I(\outTP)+I(\outTN)) \in [0,1]$ encodes the relative importance of true positives over true negatives, and $\TileB(I) = I(\outFN)/(I(\outFN)+I(\outFP)) \in [0,1]$ encodes the relative importance of false negatives over false positives. For example, $\iou$ and $\fone$ are two ranking scores corresponding to different importance values but that nevertheless lead to the same rankings as these scores are related by a monotonous increasing relationship ($\fone=\frac{2 \iou}{1+\iou} \rightarrow \frac{\partial \fone}{\partial \iou}>0$). 

\begin{figure}[tb]
\begin{centering}
\includegraphics[width=0.9\linewidth]{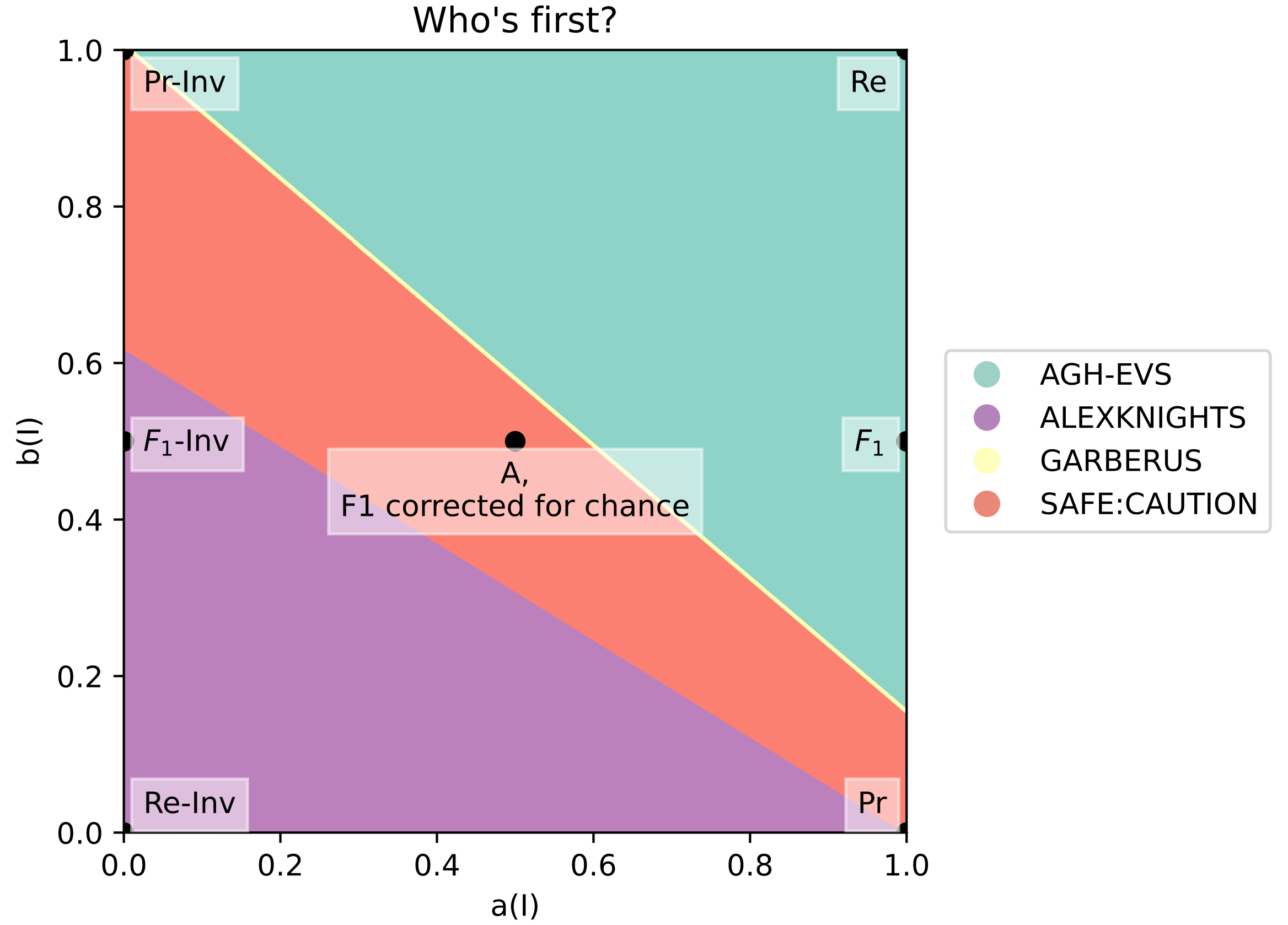}
\par\end{centering}
\caption{%
    Entity Tile, showing the best method for each $(a,b)$ pair, for the 10 methods of the IWDD challenge (taken from~\cite{Bouwmans2026Illegal}).
    \label{fig:IWDD-tile}
}
\end{figure}

\paragraph{Entity Tiles}
Following \cite{Halin2024AHitchhikers-arxiv}, the point-wise comparison of the Value Tiles obtained in the previous step can be used to obtain an \emph{Entity Tile} showing for all application-related preferences which method is the best. As a concrete example, \citet{Bouwmans2026Illegal} applied the ``Who's first?'' Entity Tile to the ten teams of the IWDD 2026 contest (see \cref{fig:IWDD-tile}).

\paragraph{Taking the chance into account}
In \cref{sec:pipeline:importance}, we already explained how we can take the chance into account for the point of view of \emph{values} by hatching areas on the Value Tiles. It is also straightforward to take the chance into account for the point of view of \emph{ranks}: it suffices to include, in the ranking, the no-skill classifiers predicting always the negative or the positive class. The former can be ranked first in the lower-left part of the Tile, meaning that there is no available method that performs better than the no-skills for the importances in that zone. Similarly, the latter can be ranked first in the upper-right part of the Tile. The part of the Entity Tile in which these no-skill classifiers do not appear first corresponds to the applicative preferences for which we have solutions.

\subsection{Guidelines}

As seen in the previous section, ranking is not just a question of scoring. In fact, we need a whole, coherent pipeline to evaluate and compare methods. Hereafter, we share some insights, expressed as numbered guidelines, on the evaluation procedure itself. 

\textbf{(G1) Compare comparable methods, that is, methods applied to the same event set.}
Rankings are meaningful only when classifiers are evaluated under identical conditions ---same task, same events, same test set, same prior distribution, same annotation protocol, \etc; this is the purpose of defining scenarios as discussed in \cref{sec:scenarios}.
Changing the terms of a scenario produces a different ``performance space''.

\textbf{(G2) Report full confusion matrices, not just score values.}
One notable strength of the \CDNET benchmarks~\cite{Goyette2014ANovel,Wang2014CDnet} is that it provides the complete confusion matrix counts of \TP, \TN, \FP, and \FN, for every evaluated method and for every video source and category. 
Most event-detection papers, by contrast, report only precision, recall, and \fone, which are \emph{TN-free} scores. 
In the Tile representation, these TN-free scores all occupy the right-hand vertical border ($\TileA=1$). 
Any score involving \TN, such as \accuracy, \tnr, \npv, or any Tile point with $\TileA<1$ requires knowing the number of true negatives, a quantity that is well-defined only when the negative class is exhaustively annotated. 
When methods are evaluated on new data outside a controlled benchmark, $\TN$s are rarely annotated to the same level of completeness, invalidating a large portion of the Tile domain. Challenge and benchmark organizers should therefore strive to define the negative class exhaustively and to report full confusion matrices so that the complete Tile domain remains accessible for downstream analyses. 
Instead of relying on a small subset of scores (typically two or three), we propose to use the Tile~\cite{Pierard2024TheTile-arxiv} and its flavors~\cite{Halin2024AHitchhikers-arxiv} as constitutive elements of our pipeline.

\textbf{(G3) Ecological argument for multi-criterion evaluation.} 
There are two points of view depending on whether an analysis targets the efficiency of a method or ranking. 

\emph{The point of view of values.}
A single metric such as $\fone$ cannot reveal asymmetric trade-offs as discussed in~\cite{Pierard2026What}.
A method that excels for positive-class detection ($\fone^+$ high) may deliberately sacrifice the negative-class ($\fone^-$ low).
The Tile exposes all such trade-offs simultaneously, acting as an ecological summary of a method's behavior across all possible application preferences.

\emph{The point of view of ranks.}
The common practice is to rank entities based on an arbitrary score. Even when the chosen score is known to lead to meaningful rankings and has clear associated applicative preferences, an important issue remains: there is a strong temptation to forget methods that are not top-ranked. However, these entities may have advantages that make them ideal for certain applications and user preferences. In contrast with the use of a single score to rank, the Entity Tile presents the rankings for a large panel of preferences so that it can consider as valuable several methods associated with diversified, yet good, performances.

\textbf{(G4) Do not confuse the point of view of values with the point of view of ranks.}
It is in general very useful to make the distinction between the point of view of values and the point of view of ranks. Let us illustrate this with two examples. Example~(1): From a \emph{value} standpoint, \fone and \iou are different scores. However, from a \emph{ranking} standpoint, they produce strictly identical orderings on any set of classifiers; they occupy the same position on the Tile. Example~(2): For \CDNET~2014~\cite{Goyette2014ANovel}, the authors observed that their ranking is well correlated with $\fone$. %
From the \emph{value} standpoint, this score is a compromise between \precision and \recall since it is a mean (harmonic) of them. But from a \emph{ranking} standpoint, \fone is not universally appropriate. Most often, another \fbeta is more appropriate. More precisely, as \citet{Pierard2026What} showed, the optimal $\beta$ minimizing rank disagreement between \precision and \recall can differ substantially from the value $\beta=1$.

\subsection{Note to Challenge Organizers}

\textbf{(N1) Avoid the attractor effect.}
When every competing method is optimized for the same single metric (typically $\fone$), the domain is progressively pulled toward a narrow region of performance space.
This homogenization suppresses diversity and stifles innovation for applications with different score preferences. Let us illustrate this point with an example. The same face-matching method, applied to the same camera feed, calls for opposite operating points depending on the scenario. Consider a face-recognition system matching faces captured by surveillance cameras against a gallery of enrolled identities. In \textit{watch-list screening}, for example, during scanning an airport crowd for people of interest, the score to minimize is the false negative: a missed match lets the target walk through unnoticed, whereas a false alarm merely costs an operator a few seconds of visual verification. In \textit{biometric access control}, for which the same matcher decides whether to unlock a secured area, the score to minimize is the false positive: a false match grants entry to an impostor, whereas a false non-match only inconveniences a legitimate user, who tries again. 
One good way to avoid this effect is to diversify evaluation scores, for instance, by adopting the Tile as a reporting standard.

\textbf{(N2) Generalization to unseen data is still an open problem.}
Keep in mind that, even with a perfect evaluation toolbox, predicting how a method will perform on unseen, potentially out-of-distribution data from benchmarking scores alone remains unsolved~\cite{Pierard2025AMethodology}.
Distribution shift ---the change in priors between the training and deployment environments~\cite{MorenoTorres2012AUnifying}--- can drastically alter rankings.
The IWDD contest~\cite{Bouwmans2026Illegal} provides a vivid illustration: methods that scored above $0.9$ on the training set dropped to below $0.6$ on the private test set collected in different scenarios, revealing strong overfitting to the training distribution.

%% file: figs/pipeline.tex
\begin{figure*}
    \centering
    \includegraphics[width=\linewidth]{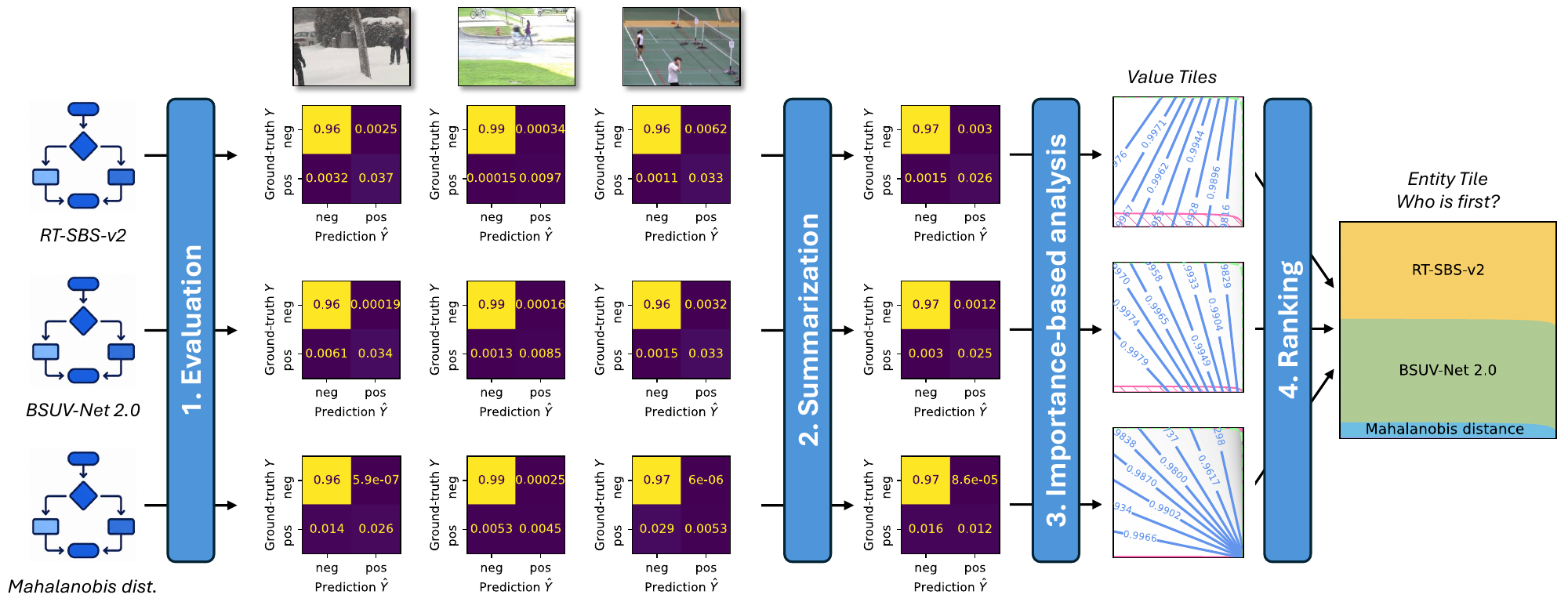}
    \caption{Illustration of a complete pipeline for evaluating and comparing the performance of methods, applied to the task of background subtraction on \CDNET. 
    The pipeline consists of four steps: 
    (1) \textit{evaluation}: this step involves determining, for each method ---as illustrated for the three methods RT-SBS-v2~\cite{Cioppa2020RealTime}, BSUV-Net 2.0~\cite{Tezcan2021BSUVNet2}, and the Mahalanobis distance~\cite{Benezeth2010Comparative}--- the confusion matrix for each video (see the first three columns);
    (2) \textit{summarization}, which involves aggregating the confusion matrices into a mixture; 
    (3) \textit{analysis of various scores} based on application-related importances, represented graphically using Value Tiles; 
    and (4) \textit{comparison} of the methods using the Entity Tile.    
    More information about Tiles can be found in \cref{fig:tile}.
    }
    \label{fig:pipeline}
\end{figure*}

%% file: figs/tile.tex
\begin{figure}[tb]
\begin{centering}
\includegraphics[width=\linewidth]{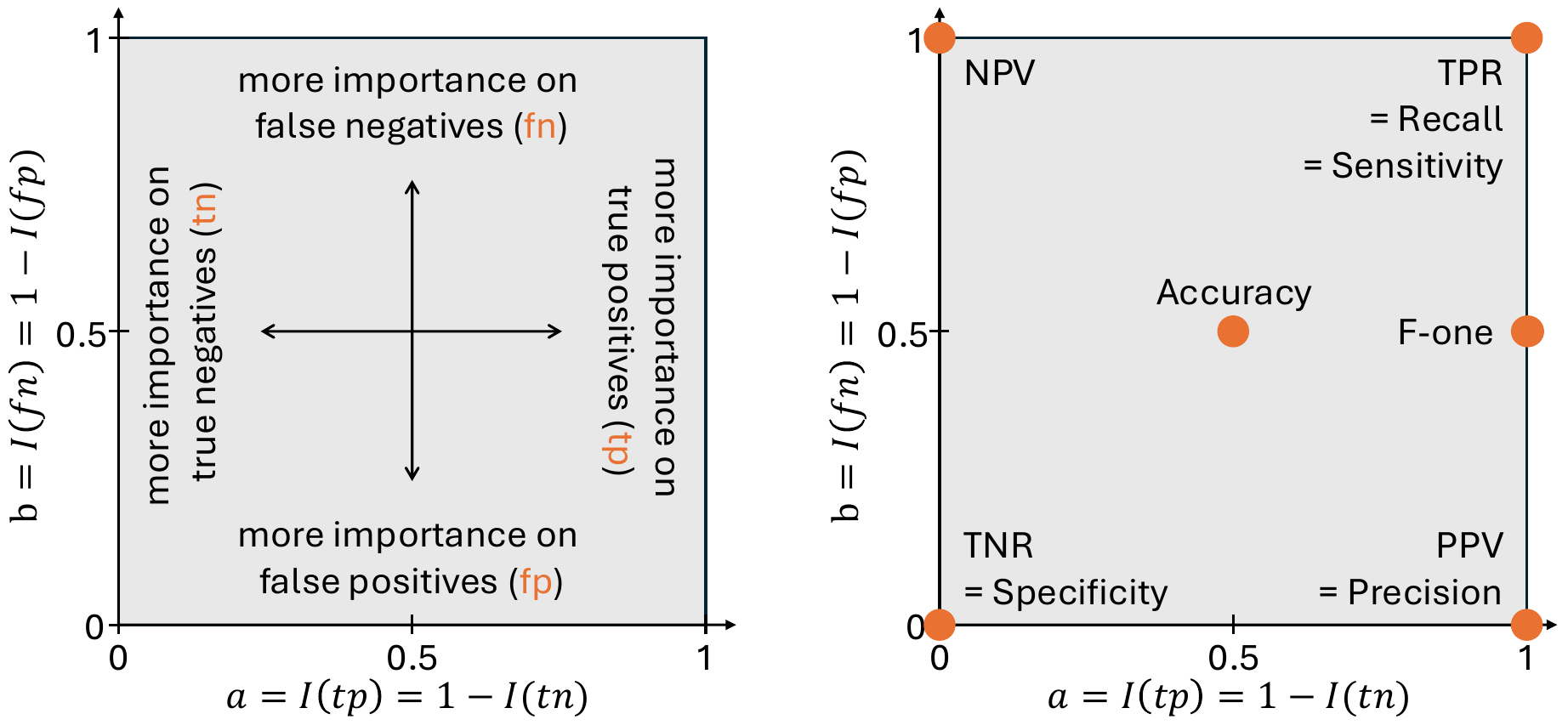}
\par\end{centering}
\caption{%
    Two interpretative readings of the Tile: a map of application-specific importances (left)
    and a map of scores to induce meaningful performance-based rankings (right)
    (taken from~\cite{Pierard2025AMethodology}).
    \label{fig:tile}
}
\end{figure}

%% file: sec/5_scenarios.tex
\section{Toward the definition of application scenarios}
\label{sec:scenarios}

When developing a method, it is important to rigorously disclose the practical choices considered under different application \emph{scenarios}. Such a description clarifies the intended use case of the method and makes comparisons with other methods more meaningful. 
It also makes the development and evaluation conditions explicit. In a practical development context, it specifies which data, prior knowledge, and operational constraints were considered. In an industrial context, it helps determine which results can be compared and under which conditions the method has been designed for. In a benchmark or challenge context, it defines the guidelines that participants must follow, including what information is available during development and evaluation. Ultimately, these application scenarios will contribute to the reproducibility of results, with the hope of limiting the uncertainty in the results to the degree of stochasticity present in most methods. Finally, there may also be requirements to document the development process. For example, the EU’s Artificial Intelligence Act regulation~\cite{EUAIAct2024} imposes a legal obligation to document the conditions under which an AI-based detection system is developed, but only if that system is considered \textit{high-risk} within the meaning of Article 6. For these reasons, scenarios need to be public and explicit, and they could be made available on a dedicated website or in an official document so that they can be referenced directly and avoid ambiguities caused by partial descriptions of development and evaluation conditions.

In this section, we define the relevant information needed to define an application \emph{scenario}. A \emph{scenario} defines the complete framework for implementing methods so that they can be compared fairly with one another by describing various \emph{characteristics} related to the \textit{data} that can be used, prior knowledge that \textit{models} can have learned, or the considered \textit{evaluation} protocol. 
In simpler terms, a scenario is a recipe, and the characteristics are the ingredients. Scenarios should be unambiguous, self-descriptive, and can be formed by different combinations of characteristics. It is therefore easy to scale the number of scenarios over time since new scenarios do not interfere with previous ones. 
Here is an example scenario. \emph{Scenario 1 defines that, during its development, a method can have access to any external data, as well as external prior knowledge about how the method works (for example, on a test set). During evaluation, methods have access to all the video frames (which means that the method is not required to be causal).} This scenario can be seen as an unrestricted scenario, aiming for state-of-the-art (SOTA) results on the proposed benchmark. It is also possible to define scenarios that share many characteristics, but that would differ for one of them. Two scenarios that share some characteristics, but not all, are the following: \emph{Scenario 2.1: 'Developers of methods can only have access to the data associated to the benchmark to develop their method without using any prior knowledge. During evaluation, methods can use all the video frames'} and \emph{Scenario 2.2: 'Developers of methods can only have access to the benchmarking data to develop their method without using any prior knowledge. During evaluation, methods have access to the video frames in a causal way only'.} With this formulation, Scenarios 2.1 and 2.2 only differ in their evaluation protocol.

In the following paragraphs, we provide a non-exhaustive list of characteristics that can be specified when defining a scenario. It can be seen as a first check-list of ingredients for setting up a scenario. 

\subsection{Data}
The data characteristics describe which information is available before and during method development. Data characteristics mainly relate on the raw data and annotations. 

\noindent \textbf{Raw data characteristics.} 
They may include the type of input modality, such as RGB, infrared, depth, audio, event cameras, metadata, or any multimodal combination of these sources. A scenario should also specify whether the data are real, synthetic, or mixed; whether they come from a single scene, multiple scenes, multiple cameras, or multiple domains; the minimum and maximum spatial resolution; the frame rate; and whether temporal information is available as isolated frames, short clips, full videos, or synchronized streams. 

\noindent \textbf{Annotation characteristics.} 
Another important distinction concerns the annotation level: data can be unlabeled, partially labeled, or exhaustively labeled at the pixel, object, frame, clip, or event level. A scenario can further describe whether labels are available only for all videos, for a subset of frames, for the first frame of each video, for bounding boxes only, for segmentation masks, for instance identities, or for higher-level event categories. Finally, the scenario should make explicit whether external data are allowed, whether pre-existing public or private datasets can be used, whether synthetic data are permitted, and whether other metadata can be exploited.

\subsection{Methods}
The method characteristics describe what kind of prior information a developer of methods is allowed to use. A method may be developed/trained from scratch on the benchmark development data only, initialized from generic pre-trained models, fine-tuned from task-specific models, or built from handcrafted rules without learning. Some scenarios may restrict the method to a fixed model after development, whereas others may allow the method to update itself during inference as more frames become available. The scenario can also specify whether ensemble methods, human-in-the-loop corrections, manual parameter tuning on validation videos, or task-specific post-processing are accepted. These elements are essential because two methods can process the same test video while relying on very different amounts of prior knowledge.

\subsection{Evaluation}
The evaluation characteristics describe the details of the event detection task, how it is evaluated and measured, and the general context of the evaluation. 

\noindent \textbf{Task.} As discussed in \cref{sec:introduction}, we cast an event detection task as a crisp two-class classification task. From an application perspective, however, a scenario should  define the prediction unit (granularity) of an event (which depends on the exact definition of the task): pixel, object, instance, frame, temporal segment, clip, or trajectory. For each unit, the positive, and negative classes must be defined explicitly so that true positives, true negatives, false positives, and false negatives have an unambiguous meaning.  

\noindent \textbf{Evaluation.} 
A series of evaluation criteria are presented in \cref{sec:performance} that can be roughly categorized as task- or hardware-related. For task-related measurements, the evaluation part of a scenario should specify the matching rules, temporal tolerance, spatial overlap threshold, treatment of ignored regions, handling of ambiguous labels, class imbalance, aggregation over videos or categories, and how rankings are computed, following \cref{sec:pipeline:ranking}. Also note that measuring true negatives may require exhaustive annotation of the absence of events, which is not always possible. For hardware-related characteristics, they could be considered as post-hoc measurements or be constitutive of development limitations. For example, one could be asked to provide the number of needed \flops or be asked, in a scenario, not to exceed a certain number of \flops.

\noindent \textbf{Contextual characteristics} should specify what information is available at inference time: the full video, only past and current frames in a causal setting, a sliding temporal window, a first annotated frame, sparse user-provided annotations, scene metadata, or no extra information beyond the raw stream. 
This distinction is important because the same video can define different tasks. For example, if the first frame is annotated and the goal is to propagate the annotation through the video, the scenario is closer to semi-supervised video object segmentation; if no frame is annotated, it is closer to unsupervised foreground or event detection.

\subsection{Others}
Other scenario characteristics describe operational constraints that are not captured by data, method, or evaluation choices alone. These include the hardware available to participants, such as CPU-only execution, embedded devices, GPUs, memory limits, sensor bandwidth, storage constraints, and energy consumption. 
They also include runtime constraints, for example, whether the method must operate in real time, near real time, offline, causally, or with a fixed maximum latency. A scenario may additionally specify reproducibility requirements, such as deterministic execution, open-source code, fixed random seeds, Docker images, model-size limits, or restrictions on proprietary services. 
In deployed settings, robustness and safety constraints may also matter: missing frames, corrupted streams, camera motion, adverse weather, domain shifts, privacy requirements, or the need to explain predictions to a human operator. These characteristics help distinguish a purely offline benchmark from a scenario intended to reflect practical deployment conditions.

%% file: sec/7_conclusion.tex
\section{Conclusion}
\label{sec:conclusion}
In this paper, we propose a framework for the development of new methods for event detection in videos. First, a large-scale dataset covering a large spectrum of environments and modalities have been presented providing both a public part and a private part to ensure fair comparison. Second, we have investigated a new performance evaluation scheme for a disruptive fair ranking of algorithms addressing both the performance in terms of detection and the performance in terms of deployability. 
Finally, we introduce the concept of an application scenario, which encompasses all the elements involved in the development of a method to avoid any ambiguity when comparing the application characteristics of methods.

%% file: sec/9_acknowledgment.tex
\section*{Acknowledgments}\label{sec:acknowledgments}

The work by S. Pi{\'e}rard was supported by the Walloon Region (Service Public de Wallonie Recherche, Belgium) under grant n°2010235 (ARIAC by \href{https://www.digitalwallonia.be/en/}{DIGITALWALLONIA4.AI}). 
We thank the bachelor students Tanushree Kakad and Kanishka Pradeep Patil from K.~K. Wagh Institute of Engineering Education and Research (Nashik, India) for their help in the construction of the large-scale dataset.

%% file: sec/10_appendix_data.tex
\appendix
\label{sec:appendix-data}
This appendix contains the detailed descriptions of the public datasets employed in the proposed large-scale dataset. 

\subsection{Urban Environments Datasets}
These datasets are the most common because the first events of interest captured by cameras generally occur in urban settings, such as  for traffic monitoring.
\begin{itemize}
\item \textbf{ATON (2003):} The ATON dataset \cite{Prati2003Detecting} is used for foreground and shadow detection in surveillance videos. It is relevant to urban RGB monitoring because it contains scenes where moving objects and their cast shadows must be distinguished from the background.
\item \textbf{OSU Thermal (2005):} The OSU Thermal dataset \cite{Davis2005ATwostage} contains thermal infrared video sequences for pedestrian and object analysis. It contains low-light or poor-visibility conditions, where thermal imagery may suffer from low contrast and limited appearance details.
\item \textbf{Terravic Motion IR (2005):} The Terravic Motion Infrared dataset \cite{Miezianko2005Terravic} consists of infrared video sequences containing moving objects in different thermal scenes. It is relevant for evaluating foreground detection and tracking under low-contrast thermal conditions and thermal noise.
\item \textbf{ETISEO (2007):} The ETISEO dataset \cite{Nghiem2007ETISEO} was developed for evaluating video-surveillance systems in urban environments. It contains videos involving pedestrians and vehicles captured in indoor and outdoor settings, and is useful for assessing detection performance under illumination changes, dynamic backgrounds, and scene-level variations.
\item \textbf{UCSD (2008):} The UCSD dataset \cite{Chan2008Modeling} contains surveillance videos of pedestrian scenes and is commonly used for anomaly detection, foreground detection, and motion analysis. It includes crowded pedestrian movement, perspective changes, camera-related variations, and occlusion.
\item \textbf{SZTAKI (2009):} The SZTAKI surveillance \cite{Benedek2009Change} benchmark provides video sequences and ground-truth masks for evaluating foreground and shadow detection methods. It is suitable for studying shadow-related challenges in urban monitoring scenarios.
\item \textbf{VIRAT (2011):} The VIRAT dataset \cite{Oh2011ALargescale} is a large-scale video-surveillance dataset designed for realistic human activity and event recognition. It contains outdoor scenes with people, vehicles, background clutter, and scale variations.
\item \textbf{SABS (2011):} The Stuttgart Artificial Background Subtraction dataset \cite{Brutzer2011Evaluation} is a synthetic benchmark for evaluating background subtraction methods. It is useful for analyzing dynamic backgrounds, shadows, and other foreground-detection challenges.
\item \textbf{BMC (2012):} The Background Models Challenge dataset \cite{Vacavant2012ABenchmark} provides real and synthetic video sequences for evaluating background subtraction and foreground detection algorithms. It focuses on outdoor surveillance scenarios affected by weather, illumination changes, dynamic backgrounds, and shadows.

\item \textbf{Audio-Visual Vehicle (AVV) (2012):}The Audio-Visual Vehicle (AVV) dataset \cite{Wang2012Real} is a multimodal urban traffic dataset developed as part of the IEEE OTCBVS Benchmark datasets. It contains 961 vehicle samples, each including an audio recording, an original image, and a reconstructed visual image.

\item \textbf{CITIC RGB-D (2013):} The CITIC RGB-D dataset \cite{FernandezSanchez2013Background} contains color and depth video data captured using RGB-D sensors. It is used for foreground-background segmentation and is relevant to challenges such as depth camouflage and inserted background.
\item \textbf{CDNet (2014):} \citet{Wang2014CDnet} proposed the Change Detection Benchmark dataset, commonly known as CDNet 2014, for evaluating foreground detection and background subtraction methods under diverse video conditions. The dataset includes challenging scenarios such as dynamic background, camera jitter, intermittent object motion, illumination variation, shadows, and PTZ camera motion.
\item \textbf{i-LIDS (2014):} The i-LIDS dataset \cite{Wang2014Person} consists of CCTV-based surveillance videos collected for intelligent detection systems. It includes realistic monitoring scenarios such as abandoned objects, doorway surveillance, and restricted-area monitoring, making it useful for evaluating event detection under shadows, occlusion, and complex urban conditions.
\item \textbf{SBI (2015):} The Scene Background Initialization dataset \cite{Maddalena2015Towards} is designed for evaluating methods that estimate a clean background image from video sequences. It is useful in urban surveillance because foreground objects may remain static for long periods, creating intermittent motion and background-initialization challenges.
\item \textbf{PETS (2016):} The PETS dataset \cite{Patino2016PETS} is a widely used video-surveillance benchmark designed for analyzing people, crowds, and object-level events in public scenes. It contains scenarios involving pedestrians, crowd movement, shadows, and occlusion.
\item \textbf{GTFD (2016):} The Grayscale Thermal Foreground Detection dataset \cite{Li2017Weighted} is used for foreground detection in thermal or infrared video sequences. It is relevant for evaluating object detection in low-contrast and noisy thermal imagery.
\item \textbf{SYNTHIA (2016):} The SYNTHIA dataset \cite{Ros2016TheSynthia} is a synthetic urban-scene dataset generated for semantic scene understanding. It contains synthetic images and video sequences with pixel-level annotations, and is useful for visual challenges such as illumination variation, shadows, and scene diversity.
\item \textbf{SBMnet (2017):} The Scene Background Modeling dataset \cite{Jodoin2017Extensive} provides a benchmark for background modeling and foreground detection. It includes challenging scenarios such as background motion, camera jitter, intermittent motion, clutter, illumination changes, and variations in video duration.
\item \textbf{Remote Scene IR (2017):} The Remote Scene IR dataset \cite{Yao2017Comparative} provides infrared video sequences captured from remote scenes for evaluating background subtraction methods. It includes challenges such as low contrast, video noise, dynamic background, camouflage, and varying foreground motion.
\item \textbf{SBM-RGBD (2017):} The SBM-RGBD dataset \cite{Camplani2017ABenchmarking} is an RGB-D benchmark for evaluating scene background modeling and moving-object detection methods. It contains RGB and depth video sequences and is useful for studying depth-related challenges, including camouflage between foreground objects and the background.
\item \textbf{MOTSynth (2021):} MOTSynth \cite{Fabbri2021MOTSynth} is a large-scale synthetic dataset designed for pedestrian detection, segmentation, and multi-object tracking. It contains synthetic pedestrian scenes with camera motion and crowded urban conditions.
\item \textbf{AGVS (2022):} The Airport Ground Video Surveillance dataset \cite{Zhang2022AGVS} focuses on change detection in airport-ground monitoring scenes. It contains long urban surveillance videos with moving objects, occlusion, and scale variation, making it suitable for evaluating event monitoring.

\end{itemize}

\subsection{Natural Environments Datasets}
These datasets are related to the passive monitoring of animals for environmental surveillance and provided by biologists and ethologists.
\begin{itemize}
\item \textbf{eMammal (2013):} The eMammal dataset \cite{Forrester2013emammal} is a camera-trap image archive for wildlife monitoring. It contains animal images collected from natural habitats and is suitable for studying detection under illumination variation, background clutter, and unconstrained outdoor conditions.
\item  \textbf{Caltech Camera Traps (2018):} The Caltech Camera Traps dataset \cite{Beery2018Recognition} contains wildlife images captured using camera traps placed in natural environments. It is used for animal detection and classification, and includes challenges such as illumination variation, nighttime imagery, motion blur, occlusion, and natural background clutter.
\item \textbf{CAMO-UOW (2018):} CAMO-UOW \cite{Li2018AFusion, Li2017Foreground} is used for camouflaged object detection in natural scenes. Animals or objects may visually blend with their surroundings, making object localisation and segmentation more difficult.
\end{itemize}

\subsection{Underwater Environments Datasets}
These datasets have been developed in the context of  the diffusion of cameras in underwater environments.
\begin{itemize}
\item \textbf{Aqu@theque (2007):} The Aqu@theque \cite{ElBaf2007Comparison} is an underwater RGB dataset used for underwater scene understanding and foreground--background separation. It is organized as an image-sequence dataset with pixel-level masks, making it suitable for instance segmentation tasks. The data is captured using a fixed underwater camera and contains marine organisms.
\item \textbf{Fish4Knowledge (2016):} The Fish4Knowledge \cite{Fisher2016Fish4Knowledge} is an underwater RGB dataset used for fish detection, fish monitoring, and underwater video analysis. It consists of image sequences captured using stationary underwater observatory cameras.  It includes sequences representing dynamic backgrounds, complex backgrounds, crowded underwater scenes, illumination changes, and varying environmental conditions, making it relevant for evaluating detection and scene analysis methods in underwater environments.
\item \textbf{Realworld Underwater Image Enhancement (RUIE) (2020):} The RUIE \cite{Liu2020RealWorld} is an underwater RGB image-enhancement benchmark. Unlike object detection or segmentation datasets, it does not provide object annotations. It is used for underwater image enhancement and restoration, particularly under marine effects.
\end{itemize}

\subsection{Maritime Environments Datasets}
\begin{itemize}
\item \textbf{MarDCT (2015):} The MarDCT \cite{Bloisi2015ARGOSVenice} is a maritime multimodal dataset designed for multimodal maritime perception, object detection, and tracking. It contains paired multimodal data captured using moving RGB and thermal sensors on maritime platforms. The dataset provides detection annotations and includes ships, boats, maritime targets, small vessels, and other small maritime objects.
\item \textbf{Maritime Obstacle Detection Dataset (MODD) (2016):} The MODD \cite{Kristan2016Fast} is an RGB dataset designed for maritime obstacle detection in autonomous navigation. It is captured from a moving unmanned surface vehicle in realistic sea environments. The dataset supports semantic segmentation through pixel-level masks and includes classes such as water, sky, and obstacles.
\item \textbf{Singapore Maritime Dataset (SMD) (2017):} The SMD dataset \cite{Prasad2017Video} is an RGB dataset used for maritime object detection under different conditions. It provides XML annotations for bounding-box object detection and includes maritime objects such as ships, boats, and buoys. The dataset contains different camera settings, including onboard moving-camera sequences, onshore stationary-camera sequences, and NIR sequences. 
\item \textbf{Ships Dataset (2018):} The Ships dataset \cite{Shao2018SeaShips} is an RGB image classification dataset used for ship recognition and classification in maritime scenes. The data is captured from a fixed camera or satellite viewpoint and consists of images labeled using a folder-based structure. The object classes include ship and non-ship categories.
\item \textbf{MASATI (2018):} The MASATI \cite{Gallego2018Automatic} is an RGB remote-sensing dataset used for maritime ship recognition. It is organized as an image-classification dataset with folder-based class labels. The images are captured from a fixed satellite or aerial viewpoint and include classes such as ship, water, and non-ship.
\item \textbf{Maritime Synthetic Dataset (2022):} The Maritime Synthetic dataset \cite{Ribeiro2022RealTime} is a synthetic dataset used for maritime object detection and tracking under controlled sea conditions. It contains synthetic images or video sequences generated using a simulated camera in a synthetic maritime environment. The object classes include ships, boats, and maritime targets.
\item \textbf{MassMIND (2023):} The MassMIND \cite{Nirgudkar2023MassMIND} is a thermal-infrared maritime dataset used for infrared scene understanding and obstacle detection. It consists of thermal images with pixel-level segmentation masks and supports instance segmentation. The dataset includes object classes such as sky, water, obstacle, living obstacle, bridge, self, and background.
\item \textbf{M3FD\_Fusion (2025):} The M3FD\_Fusion dataset \cite{Sun2025ModalityAware} is a multimodal dataset used for RGB-infrared image fusion and object detection. It consists of paired RGB-IR images captured using a moving camera with an RGB and thermal sensor system. The annotations are provided in XML/TXT format for bounding-box object detection.
\end{itemize}